\documentclass{amsart}
\usepackage[dvips]{graphicx,psfrag}
\graphicspath{ {.} }
\setkeys{Gin}{width=\linewidth,keepaspectratio=true}
\usepackage{color}
\usepackage{url}
\usepackage{float}
\usepackage{epsfig}
\usepackage{algorithm}
\usepackage{algorithmic}
\usepackage{longtable}
\usepackage{subfigure}

\usepackage{natbib}
\bibpunct{(}{)}{;}{a}{,}{;}
\usepackage[utf8]{inputenc}
\usepackage[T1]{fontenc}
\usepackage[english]{babel}

\usepackage{geometry}
\usepackage[figuresright]{rotating}

\usepackage{url}
\usepackage{algorithmic}
\usepackage{algorithm}

\usepackage{amsmath,amsthm}

\theoremstyle{definition}

\theoremstyle{remark}

\newcommand{\CommentAlg}[1]{\(\triangleright\) #1}
\usepackage{comment}
\usepackage{subfig}
\usepackage{eqnarray}
\usepackage{xcolor}
\usepackage{amssymb}

\begin{document}
\title[BRKGA for VRP with Occasional Drivers]
{A Biased Random-Key Genetic Algorithm with Variable Mutants to solve a Vehicle Routing Problem}

\author[P. Festa]{P. Festa}
\address[Paola Festa]
{Department of Mathematics and Applications, University of Napoli Federico II, Napoli 80126, Italy}
\email[P. Festa]{paola.festa@unina.it}

\author[F. Guerriero]{F. Guerriero}
\address[Francesca Guerriero]{Department of Mechanical, Energy and 
Management Engineering, University of Calabria, Rende (CS) 87036, Italy}
\email[F. Guerriero]{francesca.guerriero@unical.it}

\author[M.G.C. Resende]{M.G.C. Resende}
\address[Mauricio G.C. Resende]{Industrial and Systems Engineering,,
Univeristy of Washington,
3900 E Stevens Way NE,
Seattle, WA 98195 USA.}
\email[M.G.C. Resende]{mgcr@uw.edu}

\author[E. Scalzo]
{E. Scalzo}
\address[Edoardo Scalzo]
{Department of Mechanical, Energy and Management Engineering, University of Calabria, Rende (CS) 87036, Italy}
\email[E. Scalzo]{edoardo.scalzo@unical.it}

\begin{abstract}
The paper explores the Biased Random-Key Genetic Algorithm (BRKGA)
in the domain of logistics and vehicle routing. Specifically, the
application of the algorithm is contextualized within the framework
of the Vehicle Routing Problem with Occasional Drivers and Time
Window (VRPODTW) that represents a critical challenge in contemporary
delivery systems. Within this context, BRKGA emerges as an innovative
solution approach to optimize routing plans, balancing cost-efficiency
with operational constraints.
This research introduces a new BRKGA, characterized by a variable
mutant population which can vary from generation to generation,
named BRKGA-VM. This novel variant was tested to solve a VRPODTW.
For this purpose, an innovative specific decoder procedure was
proposed and implemented. Furthermore, a hybridization of the
algorithm with a Variable Neighborhood Descent (VND) algorithm has
also been considered, showing an improvement of problem-solving
capabilities.
Computational results show a better performances in term of
effectiveness over a previous version of BRKGA, denoted as MP.
The improved performance of BRKGA-VM is evident from its ability to optimize solutions across a wide range of scenarios, with significant improvements observed for each type of instance considered.
The analysis also reveals that VM achieves preset goals more quickly
compared to MP, thanks to the increased variability induced in the
mutant population which facilitates the exploration of new regions
of the solution space. Furthermore, the integration of VND has shown
an additional positive impact on the quality of the solutions found.
\end{abstract}

\date{April 26, 2024.}
\thanks{Technical Report.}

\maketitle
\section{Introduction}\label{sec:Introduction}
E-commerce offers significant opportunities for growth and expansion of different types of \textcolor{black}{businesses}. Indeed, online shopping opens up a number of benefits, wishes, and objectives for consumers and physical stores, that would not otherwise be possible. Online shoppers can receive their purchases quickly and conveniently at home.
 
\textcolor{black}{Additionally, the possibility of using increasingly sophisticated and efficient services for purchases is enhanced as the scientific community continues to refine optimization algorithms.} E-commerce, on the other hand, allows physical stores to expand beyond the confines of their immediate \textcolor{black}{region and country.}

In the context of last-mile delivery, to make some services practical, it is important to examine non-traditional delivery models.
In recent years, a new type of crowdsourcing, known as \textcolor{black}{\textit{crowd-shipping}}, has emerged. The development of mobile applications and \textcolor{black}{Internet} platforms, based on highly efficient algorithms, has also contributed to the growth of this phenomenon.

\textcolor{black}{Crowd-shipping embodies a cutting-edge approach that leverages people's underutilized logistics assets to provide logistics services.
Through the reduction of delivery vehicles, it has the potential to be environmentally friendly.}

The primary idea behind crowd-shipping is to \textcolor{black}{assign the delivery for some orders to common people}, known as 
\textcolor{black}{\textit{Occasional Drivers}} (ODs), who are willing to take some detours from their own paths in exchange for pay.
\textcolor{black}{Through apps or digital platforms, this innovative solution facilitates the creation of services based solely on on-demand work. Anyone who registers and shares their location, vehicle load capacity, and \textcolor{black}{available} times can \textcolor{black}{work as an occasional driver.} Based on the \textcolor{black}{shipment details}, these individuals can accept and complete delivery requests, \textcolor{black}{generating income}.}

\textcolor{black}{Some apps (e.g., Amazon Flex, Deliveroo and Glovo) offer the possibility to book deliveries for the same day as well as for subsequent days. This advance booking feature is particularly relevant for the efficient planning of vehicle routes, a central aspect of our study.
In this article, our objective is to provide support for a generic logistics company that needs to plan delivery routes for the following day based on the orders received by a specified time deadline. To facilitate this planning process, the company has at its disposal both company-employed couriers and occasional drivers who have already been booked for the upcoming delivery tasks.
\textcolor{black}{In the specified context, our study methodically concentrated on a variation of the classic Vehicle Routing Problem, known as the Vehicle Routing Problem with Occasional Drivers and Time Window constraints (VRPODTW).} 
This optimization problem concerns the planning of routes for a set of vehicles, which include both company vehicles and occasional drivers, to deliver parcels to a set of customers. The main objective is to minimize the total routing cost of company vehicles and the compensation for occasional drivers, while ensuring adherence to time and load capacity constraints.
To address the problem at hand, we propose and empirically evaluate two algorithms based on the \textcolor{black}{Biased Random-Key Genetic Algorithm (BRKGA) \citep{ref:opt_prob_4}.} These have been considered to \textcolor{black}{process the routes offline} based on the data collected up to a certain deadline, generating a routing plan that minimizes the total cost of deliveries while adhering to operational constraints.
The following day, both company and occasional drivers follow the planned routes to execute the deliveries.
This approach allows for accurate and optimized planning based on a complete set of data, ensuring that resources are used efficiently and that deliveries are executed promptly. \textcolor{black}{This way}, not only is the operational efficiency of logistics companies improved, but it also provides flexible earning opportunities for occasional drivers, contributing to creating a more resilient and responsive delivery ecosystem.}

\textcolor{black}{Genetic algorithms have demonstrated significant efficacy in addressing a wide range of complex optimization problems, and thanks to some of their distinctive properties, they can be made absolutely valid for real-world cases such as the aforementioned delivery company. Among the primary properties in this regard, we find scalability and parallelism.
The scalable nature of genetic algorithms allows for handling problems of increasing sizes without a \textcolor{black}{disproportional} increase in computational complexity. This aspect is crucial in many fields, especially in \textcolor{black}{operations and logistics}, where the problem size can rapidly grow with the increase in the number of variables and constraints.
Another notable advantage of this approach is \textcolor{black}{its} ability to leverage the natural parallelism inherent in the evolutionary process. In a genetic algorithm, many candidate solutions are explored in parallel through different populations, allowing for a quicker convergence towards optimal or acceptable suboptimal solutions. This characteristic is particularly advantageous in frenetic operational environments, where time is a precious resource, and the speed in finding optimal solutions can translate into tangible competitive advantages.
Beyond these two generic characteristics of genetic algorithms, two advantages closely related to the definition of the proposed BRKGA can be identified. Its intrinsic definition \textcolor{black}{in} the $n$-dimensional hypercube facilitates almost immediate integration of local search procedures and/or intensification strategies, making it a powerful and flexible tool that can be adapted to tackle a wide range of logistical and operational challenges.
The high adaptability of the BRKGA is another key advantage that will be discussed later, after introducing the new variant and the experimental results.}

The rest of the paper is organized as follows. In Sect.~\ref{sec:StateOfTheArt}, we describe the state of the art for the VRPODTW and for BRKGA, which represents a promising option to address the problem.
In Sect.~\ref{sec:VRPOD}, we give the details of a mathematical programming model for the problem under study, and in Sect.~\ref{sec:solution_approach}, we provide specifics on the metaheuristics based on BRKGA that are being considered to solve the problem. The computational study and analysis are described in Sect.~\ref{sec:ComputationalStudy}, and the conclusions are outlined in Sect.~\ref{sec:conclusions}.

\section{State of the Art}\label{sec:StateOfTheArt}
\textcolor{black}{The literature proposes  numerous interesting and significant studies that discuss the positive and negative aspects of crowd-shipping.}
\citet{ref:vrpod} consider for the first time the availability of ODs in the last-mile delivery process. ODs are occasional non-professional couriers, who choose to deliver goods in exchange for \textcolor{black}{some compensation}. Therefore, in this scenario a shipment company must choose how to utilize its own fleet of vehicles and the available ODs. The authors \textcolor{black}{propose a hybrid heuristic based on Variable Neighborhood Search (VNS) and Tabu Search}, to solve the vehicle routing problem with ODs. They analyze the effects of different compensation structures and conclude that adequate ones need to be built for the system to work.

\citet{ref:vrpodtw} 
study the VRPODTW, where time constraints for both customers and ODs are considered. They also consider multiple deliveries for ODs and a split delivery policy. They perform a series of computational tests to \textcolor{black}{show} the advantages of these two strategies in different scenarios. In \citet{ref:vrpodtw_vns}, a VNS approach is suggested for solving the vehicle routing problem with multiple deliveries. Subsequently, \citet{Pugliese2022} presents an approach that draws inspiration from VNS and incorporates various machine learning techniques to explore the most promising areas of the search space.
Other scientific contributions consider intermediate points in the VRPODTW as points closer to the delivery area than to the central company depot.

In particular, in \citet{ref:vrpodtw_tn}, the vehicles can collect parcels from \textcolor{black}{a central depot} and intermediate nodes, and deliver them to customers. The authors show the advantages of the intermediate nodes, by formulating the problem as a particular instance of two-echelon vehicle routing problem.
The potential benefits of using intermediate points to support driver deliveries are also discussed \textcolor{black}{by \citet{ref:sampaio}. In particular,} the authors highlight the benefits of using this system when pick-up and drop-off locations are distant and OD time windows are tight.
In \citet{ref:vrpodtw_od}, intermediate locations are considered as additional elements of the occasional system rather than either intermediate depots owned by shipping companies or points of interchange for couriers.
More specifically, a new scenario of occasional depots is put forth in an effort to provide other resources unused by the crowd at the service of the delivery system. These depots are used only when needed, and \textcolor{black}{as a consequence}, their owners are compensated when the occasion arises. In addition, the authors divide the set of ODs into two groups based on the tasks carried out and the compensation plan. They outline the advantages of this scenario and present a mixed integer programming model.
Other Vehicle Routing Problem with Occasional Drivers (VRPOD) variants take into account electric vehicles (see, e.g., \citet{ref:vrpod_green_1} and \citet{ref:vrpod_green_2}) or consider specific scenarios where information is dynamic or affected by uncertainty (i.e., stochastic/online VRPOD) (see, e.g., \citet{ref:vrpod_online2}, \citet{ref:vrpod_online}, \citet{DIPUGLIAPUGLIESE2023102776}).

Regarding the strategies developed to solve the VRPOD, the literature presents \textcolor{black}{several contributions} that provide high performance metaheuristics like genetic algorithm, tabu search, simulated annealing, and adaptive large neighborhood search. 
Genetic algorithms represent a significant class of algorithms that usually produce excellent results when solving combinatorial optimization problems. 
Considering the good results we obtained in our previous contribution, by using the BRKGA \citep{ref:vrp_brkga_3}, in this paper we present a new version of this approach, which considers a variation in the population generation, and we also propose a metaheuristic which combines the proposed new version with a local search strategy.

BRKGA has been firstly proposed by \citet{ref:opt_prob_1} with the goal of developing a sophisticated \textcolor{black}{general search metaheuristic that can be enhanced} with intensification and diversification strategies to address several types of problems.
The approach has been applied successfully to several combinatorial optimization problems, including the VRP (see, e.g., \citet{ref:opt_prob_2}, \citet{ref:vrp_brkga_1}, \citet{ref:vrp_brkga_2}, \citet{ref:vrp_brkga_4}, and \citet{ref:brkga_opt_prob_3}).
\textcolor{black}{See \citet{LonPesAndRes24a} for a recent review of the literature of BRKGA.}

Several variants of BRKGA have been introduced in the scientific literature. More specifically, \citet{ref:brkga_mp} propose two different versions of the crossover operator. In the first one, a crossover between two or more parents is described, while in the second one, a crossover based on the gender of chromosomes is developed. The results, collected on a set of covering and packing instances, show that the first variant outperforms the standard one, in which the crossover operator is defined exclusively between two parents. It is also possible to adapt the BRKGA in case the problem to be addressed is multi-objective (see, e.g., \citet{ref:biObj_brkga}). It was demonstrated that the independent evolution of multiple populations is another key option that can increase the possibility of finding good solutions and accelerating convergence (see, e.g., \citet{ref:brkga_multiPop_1}, \citet{ref:brkga_multiPop_2}, \citet{ref:brkga_multiPop_3}, and \citet{ref:brkga_multiPop_4}). Recently, \citet{ref:brkga_mp_ipr} propose a hybridization of the BRKGA with a prominent heuristic search technique: the path-relinking, which would result in a notable improvement in both the running time and result quality. The authors consider different versions of the path-relinking as an intensification strategy.

Regarding the contributions of this paper, we develop a new version of BRKGA which relies on the idea that the mutant population may vary from generation to generation.  Previous work on dynamic BRKGA include \citet{CHAVES2018a}, \citet{ChaLorNog2021a}, and \citet{Zambelli2022a}. In additional, to solve the VRPODTW, we propose this new variant of BRKGA and its specific hybridization with a Variable Neighborhood Descent (VND). In both approaches, we also consider an implicit path-relinking as the intensification strategy, as well as an evolution restart as the diversification strategy.

\section{The VRPODTW Description}\label{sec:VRPOD}
\textcolor{black}{In this section, the mathematical programming model of the VRPODTW proposed by \citet{ref:vrpodtw} is \textcolor{black}{reviewed}. Let $C$ be the set of customers, let $o$ be the origin node, which is a depot, common to all drivers. Let $D$ be the set of company drivers, and $b$ their destination nodes, let $K$ be the set of available occasional drivers, and $V=\{v_k\}_{k\in K}$ the set of their destination nodes. We define the node set as $N:=C\cup V\cup\{o, b\}$.
Each node pair $(i, j)\in N\times N$ is associated with a positive cost $c_{ij}$ and a travel time $t_{ij}$, which satisfy the triangle inequality. Each customer $i$ has a number of packages requested $q_i$, each company driver and occasional driver $k$ has a maximum transport capacity $Q$ and $Q_k$, respectively. Each $i\in C\cup D\cup K$ has a time window $[e_i, l_i]$.}

\textcolor{black}{
Let $x_{ij}$ be a binary variable equal to 1 if and only if a company driver moves along the arc $(i,j)$, and let $r_{ij}^k$ be a binary variable equal to 1 if and only if occasional driver $k$ moves along the arc $(i,j)$. Let $y_i$ and $w_i^k$ be the available capacities of company driver and occasional driver, after delivering to $i\in C$, respectively. Let $s_i$ and $f_i^k$ be the arrival times of company driver and the occasional driver $k$ at customer $i$, respectively. The mathematical formulation of the problem is given below.}

\textcolor{black}{
\begin{eqnarray}
&&\min\sum_{i\in C\cup\{o\}}\sum_{j\in C\cup\{b\}} c_{ij}x_{ij}+\sum_{k\in K}\sum_{i\in C\cup\{o\}}\sum_{j\in C\cup V} \rho c_{ij}r_{ij}^k- \sum_{k\in K}\sum_{j\in C}c_{ov_k}r_{oj}^k\label{eq:01}\\
&&s.t.\nonumber\\
&&\sum_{j\in C\cup\{b\}}x_{ij}-\sum_{j\in C\cup\{o\}}x_{ji} = 0\hspace{1cm}\forall i\in C\label{eq:02}\\
&&\sum_{j\in C}x_{oj}-\sum_{j\in C}x_{jb} = 0\label{eq:02bis}\\
&&y_j \geq y_i+q_jx_{ij}-Q(1-x_{ij})\hspace{1cm}\forall i\in C\cup\{o\},\forall j\in C\cup\{b\}\label{eq:03}\\
&&s_j\geq s_i+t_{ij}x_{ij}-M(1-x_{ij})\hspace{1cm}\forall i, j\in C, M \text{ be a large number}\label{eq:05}\\
&&e_i\leq s_i\leq l_i\hspace{1cm}\forall i\in C\label{eq:06}\\
&&\sum_{j\in C}x_{oj}\leq |D|\label{eq:07}\\
&&\sum_{j\in C\cup\{v_k\}}r_{ij}^k-\sum_{j\in C\cup\{o\}}r_{ji}^k=0\hspace{1cm}\forall i\in C,\forall k\in K\label{eq:08}\\
&&\sum_{j\in C\cup\{v_k\}}r_{oj}^k-\sum_{j\in C\cup\{o\}}r_{jv_k}^k=0\hspace{1cm}\forall k\in K\label{eq:09}\\
&&\sum_{k\in K}\sum_{j\in C\cup\{v_k\}}r_{oj}^k\leq |K|\label{eq:10}\\
&&\sum_{j\in C}r_{oj}^k\leq 1\hspace{1cm}\forall k\in K\label{eq:11}
\end{eqnarray}}

\textcolor{black}{
\begin{eqnarray}
&&w_j^k\geq w_i^k+q_ir_{ij}^k-Q_k(1-r_{ij}^k)\hspace{1cm}\forall i\in C\cup\{o\},\forall k\in K,\forall j\in C\cup\{v_k\}\label{eq:12}\\
&&w_o^k\leq Q_k\hspace{1cm}\forall k\in K\label{eq:13}\\
&&f_i^k+t_{ij}r_{ij}^k-M(1-r_{ij}^k)\leq f_j^k\hspace{1cm}\forall i\in C,\forall k\in K,\forall j\in C\cup\{v_k\},M \text{ be a large number}\label{eq:14}\\
&&f_i^k\geq e_{v_k}+t_{oi}\hspace{1cm}\forall i\in C,\forall k\in K\label{eq:15}\\
&&f_{v_k}^k\leq l_{v_k}\hspace{1cm}\forall k\in K\label{eq:16}\\
&&e_i\leq f_i^k\leq l_i\hspace{1cm}\forall i\in C\label{eq:17}\\
&&\sum_{j\in C\cup\{o\}}x_{ij}+\sum_{j\in C\cup\{v_k\}}\sum_{k\in K}r_{ij}^k = 1\hspace{1cm}\forall i\in C\label{eq:18}\\
&&x_{ij}\in\{0,1\}\hspace{1cm}\forall i,j\in N\label{eq:19}
\\
&&r_{ij}^k\in\{0,1\}\hspace{1cm}\forall i,j\in N,\forall k\in K\label{eq:20}\\
&&y_i\in[0, Q]\cap\mathbb{Z}\hspace{1cm}\forall i\in C\cup\{o, b\}\label{eq:21}\\
&&w_i^k\in[0, Q_k]\cap\mathbb{Z}\hspace{1cm}\forall k\in K,\forall i\in C\cup\{o, v_k\}\label{eq:22}\\
&&f_i^k\geq 0\hspace{1cm}\forall k\in K,\forall i\in C\cup\{o, v_k\}.\label{eq:23}\\
&&s_i\geq 0\hspace{1cm}\forall i\in C.\label{eq:24}
\end{eqnarray}}

\textcolor{black}{
The goal is to minimize total costs, which include the cost of OD remuneration considering the compensation factor $\rho$ and the cost of company driver routing. We outline two separate sets of constraints. Equations (\ref{eq:02})--(\ref{eq:07}) represent constraints of classical capacitated vehicle routing problem with time windows, dedicated to the company vehicles, while (\ref{eq:08})--(\ref{eq:18}) control the presence of ODs in the delivery system. In particular, constraints (\ref{eq:02}) and (\ref{eq:02bis}) ensure the flow conservation. Equations (\ref{eq:03}) are the capacity constraints. Constraints (\ref{eq:05}) and (\ref{eq:06}) manage the arrival time and time windows constraints, respectively. The last constraint (\ref{eq:07}) of the first set establishes a maximum limit for available company vehicles. In the second set of constraints, both series of equations (\ref{eq:08}) and (\ref{eq:09}) represent flow conservation of ODs. Constraint (\ref{eq:10}) establishes a maximum limit of available ODs, while constraints (\ref{eq:11}) make sure that each OD only ever leaves their origin node once. Equations (\ref{eq:12}) are the classical capacity constraints, and constraints (\ref{eq:14})--(\ref{eq:17}) manage the time windows of all nodes. In particular, the arrival time at node \textit{j} is computed by constraints (\ref{eq:14}), while each customer is visited and each driver performs deliveries inside their respective time windows according to constraints (\ref{eq:15})--(\ref{eq:17}).
Constraints (\ref{eq:18}) ensure that each customer is served once and only once. Finally, the domains of the variables are defined by equations (\ref{eq:19})--(\ref{eq:24}).
}

\section{The Solution Approaches}\label{sec:solution_approach}
In this section, we firstly present the proposed new version of the BRKGA, that is the \textcolor{black}{Biased Random-Key Genetic Algorithm with Variable Mutants (BRKGA-VM)}. Successively, we describe a metaheuristic, in which the BRKGA-VM is integrated with three other algorithms, designed to solve the VRPODTW, that is a path-relinking, a restart strategy, and a local search.

\subsection{The Biased Random-Key Genetic Algorithm \textcolor{black}{with} Variable \textcolor{black}{Mutants}}
The traditional structure of the genetic algorithm that forms the basis of the new variant is described in what follows. The algorithm starts by creating the first generation of evolution. Each generation is composed of one or more sets of chromosomes, also known as populations, which are represented by random--key $n$--vectors.
Consequently, each chromosome undergoes decoding through a decoder, and each population is separated into two groups based on the fitness value: the elite sub-population, which contains the chromosomes with the best fitness, and the non-elite sub-population, which contains the remaining chromosomes.

A population of the next generation is derived from three different sources: the elite sub-population from the previous generation; sub-population consisting of entirely new chromosomes, named mutant chromosomes; and sub-population produced by crossover between parents of the previous generation.

In the traditional version, the size of a mutant sub-population is initially fixed and remains constant throughout the \textcolor{black}{iterations of the algorithm}.

In the proposed version, we introduce an adaptive mechanism through the use of a variable percentage of mutants. This feature allows for dynamically altering the population composition in response to current performance and the specific needs of the problem at hand.
The sizes of the sub-populations are determined as follows. The size of an entire population is calculated as $p:= \alpha\cdot n$, where $\alpha\geq 1$ is called population size factor and $n$ is the chromosome size. The size of the elite sub-population is defined as $p_e:=p\cdot pct_e$, where $pct_e\in[0.1, 0.25]$ is the fixed percentage of elite chromosome parameter. Finally, \textcolor{black}{as opposed to} the standard version of the BRKGA, the size of a mutant sub-population may vary from one generation to the next if the cost of the best current solution is not improved; in particular, it is $p_m:=p\cdot pct^i_{vm}$, where $pct^i_{vm}\in[0.1, 0.6]$ is the variable mutant percentage, as defined in equation~\ref{eq:variableMutant}.

At the start of the procedure, this mutant percentage is set to a value $pct^0_{vm}\in [0.1, 0.3]$, which is determined using the \textit{irace} automatic configuration program and remains constant throughout the \textcolor{black}{iterations of the algorithm}.  The variable mutant percentage can subsequently increase by another fixed value, $pct_{mi}\in[0, 0.3]$, if the best current solution has not improved after $\frac{h}{2}, \frac{h}{4},$ or $\frac{h}{8}$ iterations, where $h$ is the restart parameter. 

Consequently, between two consecutive restarts, the variable mutant percentage can be increased up to three times, and we can collectively define it as follows:
\begin{equation}\label{eq:variableMutant}
 pct^i_{vm}:=pct^0_{vm} + min(i \cdot pct_{mi}, 0.6-pct^0_{vm}),
\end{equation}
where $i$ is set equal to 0 at the start of the genetic algorithm and after each time the restart strategy is used; while it is set equal to $1, 2, 3$ after $\frac{h}{2}, \frac{h}{4},$ or $\frac{h}{8}$ iterations, respectively.

This variable parameter is particularly useful for several reasons:

\begin{itemize}
    \item \textcolor{black}{Exploration vs intensification. In a genetic algorithm, there is always a tension between exploration and intensification. A fixed percentage of mutants may not be \textcolor{black}{ideal}  for both objectives throughout all phases of the algorithm. Having a variable percentage allows the algorithm to adapt; it can increase the percentage of mutants when broader exploration is needed and keep it constant when it is more advantageous to focus on existing solutions.}

    \item \textcolor{black}{Avoiding stagnation. If the algorithm gets stuck in a local optimum and fails to find better solutions, increasing the percentage of mutants can introduce greater variability into the population, helping the algorithm to escape from these areas and explore new regions of the solution space.}

    \item \textcolor{black}{Adaptability to different problems. Different problems may require varying levels of exploration and intensification. A variable percentage of mutants makes the algorithm more flexible and adaptable to a variety of problems without the need for manual adjustment of this parameter.}
\end{itemize}

\textcolor{black}{
Additionally, the formula includes a safety term, represented by $0.6 - pct^0_{vm}$, which prevents the mutant percentage from becoming excessively high and surpassing permissible limits. This term also safeguards against an excessive reduction in the number of crossovers, thereby ensuring a certain level of stability in the algorithm as suggested in~\cite{ref:opt_prob_4}.}
We refer to the \textcolor{black}{BRKGA} with this variable mutant sub-population as BRKGA-VM.

\begin{table}
\caption{Summary of parameters. 
}\label{tab:parameters}
{\small
\begin{tabular*}{\hsize}{@{}@{\extracolsep{\fill}}llllll@{}}
\hline
{Operator} & & {IPR-Per\phantom{r}} & & {Others} & \\
\hline
$pct_e$ & pct. of elite chr. & $sel$ & individual selection & $\alpha$ & population size factor\\
$pct^0_{vm}$ & initial pct. of mutant variable chr. & $md$ & min. distance among chr. & {$m$} & {number of populations}\\
{$pct_{mi}$} & {pct. increase of mutant variable chr.} & {$pct_p$} & {pct. of path size} & {$\rho$} & {compensation factor}\\
{$\pi_t$} & {number of parents in the crossover} & \phantom{a} & & {$prDel$} & {probability of delivery} \\
{$\pi_e$} & {number of elite parents in the crossover} & & & {$h$} & {restart parameter}\\
{$\phi$} & {bias function for the crossover} & & &{$wi$} & max consecutive iter.\\
\hline
\end{tabular*}} 
\end{table}

To solve the VRPODTW, a multi-population and multi-parent BRKGA-VM is devised.
\textcolor{black}{The generations consist} of $m$ independent populations and a next generation is built by using the three evolutionary operators described in what follows.

The first one is the \textit{copy repeat}, i.e., all elite chromosomes of the previous generation are copied to the populations of the next generation.
With the \textit{variable-mutant} operator, new random chromosomes are introduced in each iteration; they are built using the same technique as the initial chromosomes.
The remaining part of the populations, $p(1-pct_e-pct^i_{vm})$, is generated by the \textit{multi-parent crossover}. Three parameters must be chosen to define the crossover: the total number of parents ($\pi_t$), the number of elite parents ($\pi_e$), and the probability that each parent would carry genes to their offspring. The probability is calculated taking into account the bias of the parent, which is defined by a pre-determined, non-increasing weighting bias function ($\phi$). The set of various types of functions for it is \textcolor{black}{$D_\phi$:=\{constant, logarithmic, linear, polynomial, exponential\}.}
This operator enables the transfer of genes from a combination of different chromosomes.
For a detailed description of the multi-parent crossover operator and the bias function, the reader is referred to \citet{ref:brkga_mp_ipr}.

\subsection{The Decoder}\label{sub:Decoder}
In this section, we present the details of the chromosome decoding algorithm.
Each chromosome is a vector consisting of $n=|C|+|D|+|K|$ random-keys, also referred to as genes, which are random real numbers that are taken from the interval [0,1).

Our decoder evaluates each chromosome of the current generation to build solutions for the VRPODTW problem, specifically determining the sequence in which each driver should visit customers. It also calculates the corresponding fitness, which quantifies the total cost needed to complete all deliveries.
This total cost includes both the routing expenses for company drivers and the compensations for occasional drivers. The complete details of the decoding process are provided \textcolor{black}{in  Algorithm~\ref{alg:DecoderAgthm}}.

The procedure begins (lines~\ref{lineAl:startInit}--\ref{lineAl:endInit}) by accepting a chromosome as input and initializes an empty list called \textit{paths} and a variable \textit{sol} set to zero. The list will contain the paths of all drivers, both company and occasional drivers. A path, associated with a driver $d$, is an ordered list of nodes representing the order in which $d$ visits these nodes. In line~\ref{lineAl:separation}, the \textit{separate()} function divides the input chromosome into two distinct sub-chromosomes based on the number of customers and drivers. Specifically, the first sub-chromosome contains as many genes as there are customers, while the second sub-chromosome contains genes corresponding to the number of drivers. This allows for separate manipulation of the customer and driver sets.
In the loop of line~\ref{lineAl:pathInit}, the driver paths are initialized. Given that the origin node of all drivers is the depot, then all paths are initialized as $\{s\}$.

\textcolor{black}{In  lines}~\ref{lineAl:custSort} and~\ref{lineAl:driverSort}, the algorithm sorts the genes, of both the customer and driver sub-chromosomes in non-decreasing order. This sorting mechanism fulfills two specific roles. In the customer part of the chromosome, it determines the sequence \textcolor{black}{in which} customers will be handled. In the driver part of the chromosome, it establishes a hierarchy for considering drivers for assignment, aimed at delivering to customers. This brings us to a special phase focused on each customer (line~\ref{lineAl:CustPhase}). In this phase, we check \textcolor{black}{whether it is} possible to add a detour for a specific driver to deliver to \textcolor{black}{the} customer, considering factors like delivery time and \textcolor{black}{vehicle capacity}.

\begin{algorithm}[t]
\begin{algorithmic}[1]
\STATE {\bf input} \textit{chromosome}\label{lineAl:startInit}
\STATE {\bf set} $paths =$ [ ]
\STATE {\bf set} $sol = 0$
\STATE {\bf set} $[customerChrom, driverChrom] = chromosome.separate()$\label{lineAl:separation}
\FOR{\textit{d} in $driverChrom$}\label{lineAl:pathInit}
 \STATE $paths[\textit{d}] = [s]$
\ENDFOR\label{lineAl:endInit}
\STATE $customerChrom.sort(genes)$\label{lineAl:custSort}
\STATE $driverChrom.sort(genes)$\label{lineAl:driverSort}
\STATE\CommentAlg{Start processing phase for customers}\label{lineAl:CustPhase}
\FOR{\textit{c} in $customerChrom$}
 \FOR{\textit{d} in \textit{driverChrom}}\label{lineAl:startUpdates}
 \IF{$capacityCheck$(\textit{d, c}) and $timeCheck$(\textit{d, c}) and $delivery$(\textit{prDel})\label{lineAl:checks}}
 		\STATE $paths[\textit{d}].append(c)$\label{lineAl:pathUpdate}
 		\STATE $sol$ $+= cost(c, paths[d][-1]$) $\times$ $compensation$($d$)\label{lineAl:solUpdate}
 		\STATE $TW$[\textit{d}].\textit{update}()\label{lineAl:TWUpdate}
 		\STATE $capacity$[\textit{d}].\textit{update}()\label{lineAl:capacityUpdate}
 		\STATE break
 \ENDIF
 \ENDFOR\label{lineAl:endUpdates}
 \IF{$c$ is not served}\label{lineAl:startExlud}
 \STATE $sol =$ $+\infty$
 \STATE \textit{chromosome} excluded
 \ENDIF\label{lineAl:endExlud}
\ENDFOR
\STATE $sol$ $+= solFinalization()$\label{lineAl:solFinalization}
\STATE {\bf return} $paths$, $sol$
\end{algorithmic}
\caption{VRPODTW decoder}\label{alg:DecoderAgthm}
\end{algorithm}

For each customer $c$, this processing phase ends when a driver is found to serve them.
In particular, in lines~\ref{lineAl:startUpdates}--\ref{lineAl:endUpdates}, the paths and the fitness are built and computed, respectively.
Given a customer $c$ and a driver $d$, the procedure checks, with the \textit{capacityCheck()} function, if the capacity of $d$ is sufficient to transport the packages requested by $c$. The procedure also checks, with \textit{timeCheck()}, if the customer's time window is compatible with the travel time from the last node visited by $d$, and if it is compatible with the time window of the driver itself.
These checks are necessary because we must ensure that the customer is served at the time indicated and that the driver makes deliveries within their respective time windows.
It should be emphasized that \textcolor{black}{these checks} are designed to guarantee the feasibility of the generated solutions. However, an exception arises when the ordering dictated by the genes within the chromosomes results in the exclusion of at least one customer from being served. Under such circumstances, the algorithm deems the solution infeasible and consequently removes the corresponding chromosome from the population (lines~\ref{lineAl:startExlud}--~\ref{lineAl:endExlud}). \textcolor{black}{This is not a standard action implemented in previous BRKGAs.} The BRKGA then proceeds to evaluate the next chromosome in the sequence.

In line~\ref{lineAl:solUpdate}, the solution is updated by adding the cost to go from the last node previously visited to the processed customer \textit{c}. The cost depends on the compensation rate, \textcolor{black}{which is 1 for company drivers and less than 1 for occasional drivers}.
Before concluding the \textit{for} loop and processing the next customer, we \textcolor{black}{must update} the transport capacity and the time window \textcolor{black}{of driver} $d$ (lines~\ref{lineAl:TWUpdate} and~\ref{lineAl:capacityUpdate}).
After completing the customer processing phase, the solution cost is finalized in line~\ref{lineAl:solFinalization}. At this point, it becomes possible to accurately calculate the cost for each driver to travel from their last served customer to their destination. This cost is then added to the overall solution. 
In fact, at the same time, the cost of the driver's original route, which is from the depot to their destination, is subtracted. This update is made to align with the compensation scheme for occasional drivers (see Eq.(\ref{eq:01})).
The decoder at the end of the execution returns the path of each driver and the fitness, i.e., the solution cost.

\begin{figure}
 \centerline{\includegraphics[scale=0.7]{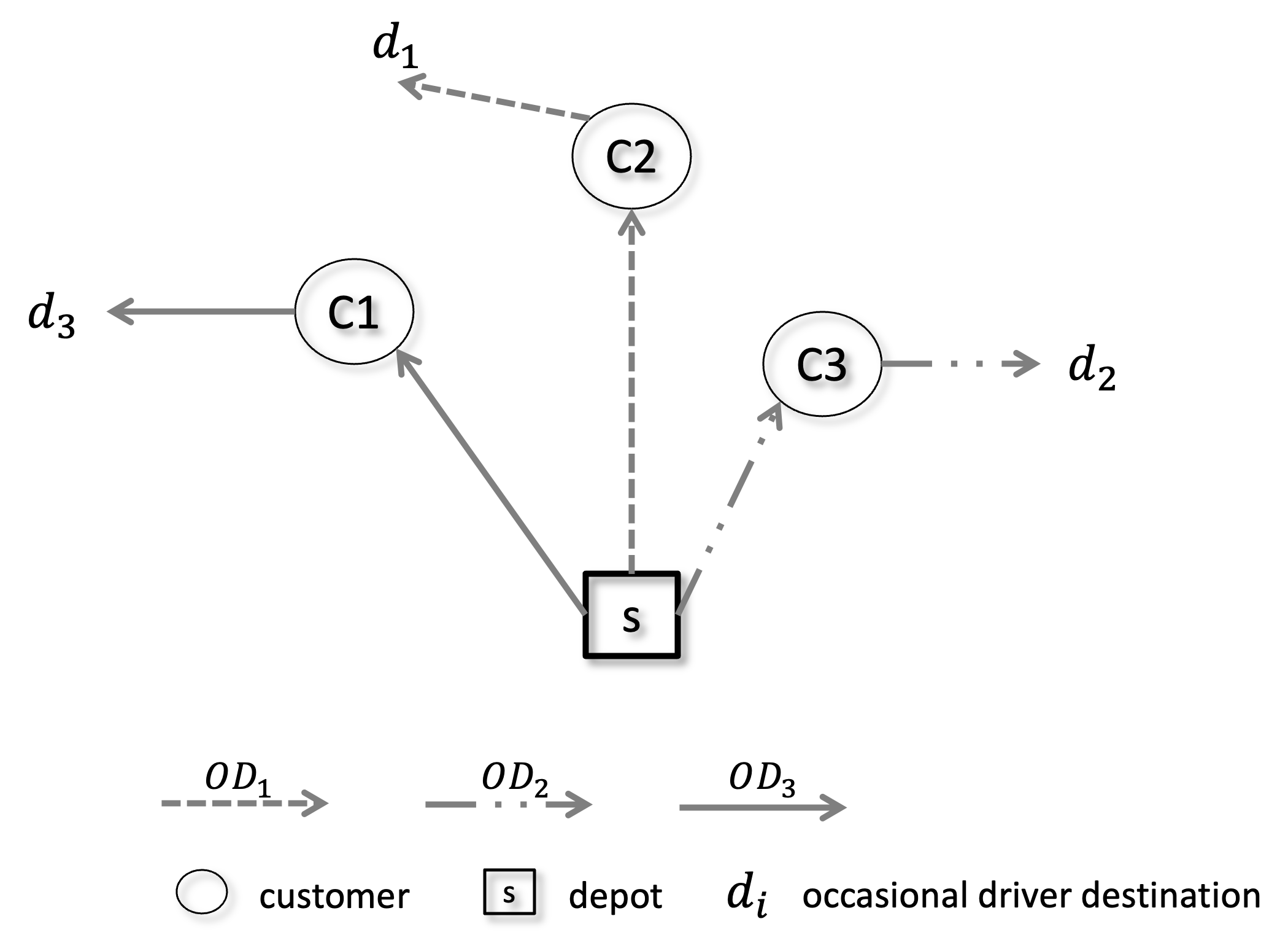}}
\caption{\textrm{\textcolor{black}{Graphical representation of an optimal solution in a scenario specifically designed to highlight the importance of the random component in the decoder. Note that each occasional driver has a capacity of 3, while each customer has a demand of 1. The optimal solution, with a cost of 7.8, was identified by both CPLEX and BRKGA using the random component of the decoder.}}}\label{fig:prDel_1}
\end{figure}

\begin{figure}
 \centerline{\includegraphics[scale=0.7]{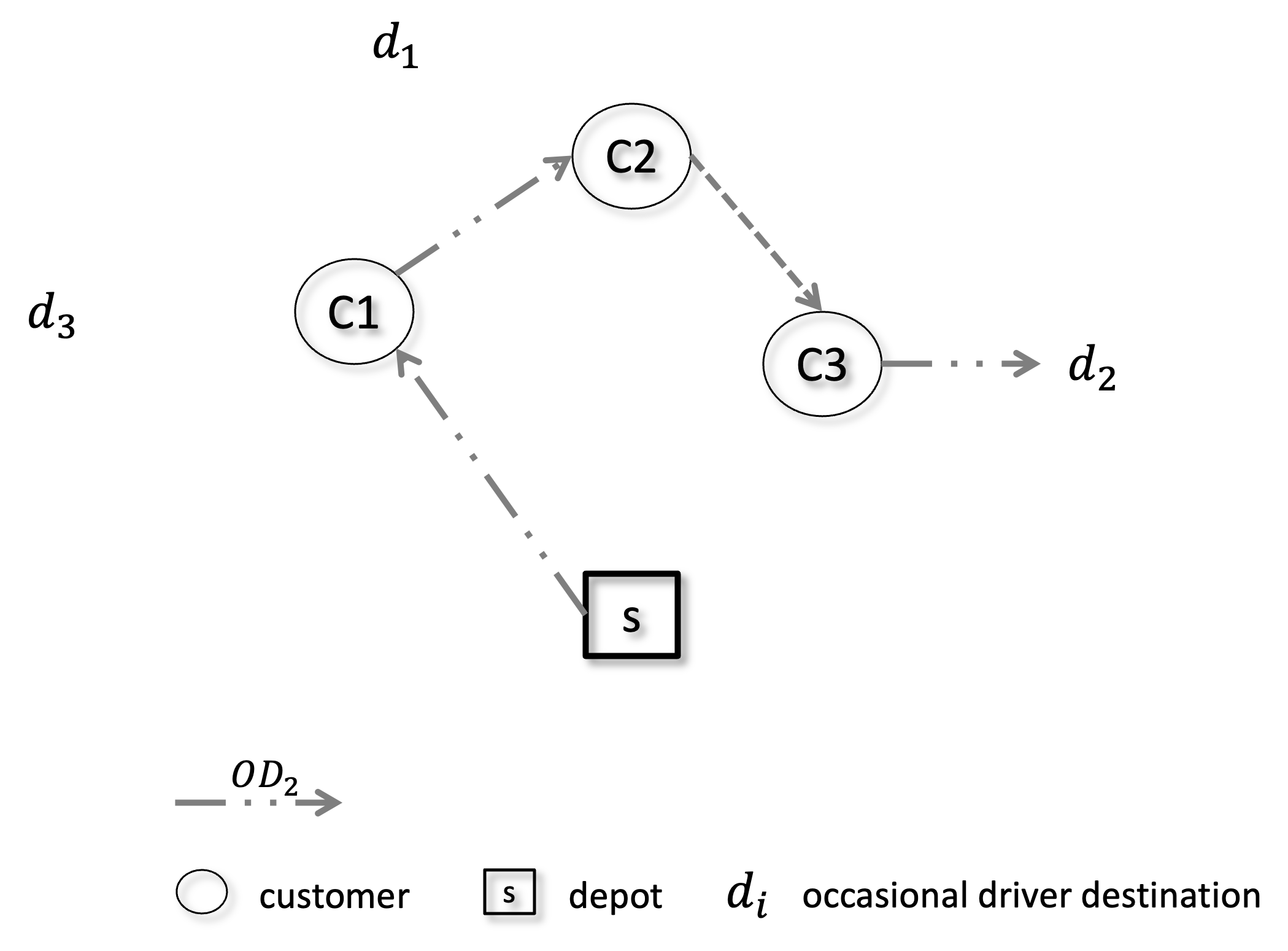}}
\caption{\textrm{\textcolor{black}{Graphic representation of the solution provided by BRKGA without the random component in the same scenario as Figure~\ref{fig:prDel_1}. The cost of the identified solution is 19.2.}}}\label{fig:prDel_2}
\end{figure}

\textcolor{black}{As mentioned previously, in the delivery assignment process, drivers and customers are chosen based on the order set in the sub-chromosomes. However, if \textcolor{black}{we do not} include a random element in the driver selection, we risk introducing a bias, consistently favoring the same drivers who appear first. To mitigate this effect, each driver is selected based on a specific probability, denoted as \textit{prDel}.
The need for this random element becomes apparent when considering an example where an optimal solution requires selecting all available drivers, with a few deliveries assigned to each (Figure~\ref{fig:prDel_1}). In addition, the time windows of customers and drivers are assumed to be set in such a way as to offer great flexibility in the sequence of service to customers. In other words, the time windows are large enough to allow any combination of orders in serving customers without violating time constraints. In this scenario, if no random element is introduced, the decoder will always choose the first driver induced by the chromosome until the transport capacities are exhausted (Figure~\ref{fig:prDel_2}).
Therefore to overcome this problem, we introduce a random component via the \textit{delivery()} function (line~\ref{lineAl:checks}), as delineated in Algorithm~\ref{alg:DeliveryFunction}.
}

\textcolor{black}{Specifically, this function employs a random value, denoted as \textit{RAND} that is compared against the predefined parameter \textit{prDel}, which represents the probability of a particular delivery occurring. If both capacity and time check functions are not violated and \textit{RAND} is less than \textit{prDel}, the algorithm proceeds to include the customer \textit{c} at the end of the path of the driver \textit{d}, as indicated in line~\ref{lineAl:pathUpdate} \textcolor{black}{of Algorithm~\ref{alg:DecoderAgthm}}.
To ensure the uniqueness of the chromosome decoding process and to ensure that its fitness is kept the same across different generations (\cite{ref:opt_prob_4}), the \textit{RAND} value is calculated in a way that depends on the specific sequence of genes in the chromosome. Specifically, the genetic sequence of the processed chromosome is used to calculate a seed (line~\ref{lineAl:seed}), which is subsequently used to determine the \textit{RAND} value (line~\ref{lineAl:Rand}).
The function, named \textit{SeedGen()}, assign to each chromosome $chr:=[gene_1,\ldots, gene_n]$ a specific value \textcolor{black}{given by}
$$\textit{SeedGen(chr)} := \sum_{i=1}^{n} \left( 2^{n-i} \times \left\lfloor 100 \times \textit{gene}_i \right\rfloor \right).$$
This methodology, by exponentially increasing the weight of genes based on their position, guarantees a unique variety in each chromosomal sequence, thus reducing the possibility that two distinct chromosomes end up having the same output.
}

\begin{algorithm}[t]
\begin{algorithmic}[1]
\STATE {\bf input} \textit{prDel, chromosome}
\STATE {\bf set} $seed = 0$
\STATE\CommentAlg{Start \textit{SeedGen} function}
\FOR{\textit{gene} in \textit{chromosome}}
  \STATE $seed = seed \times 2 + \text{int}(gene \times 100)$\label{lineAl:seed}
\ENDFOR
\STATE $RAND$ = \textit{randomValue}\textit{($seed$)}\label{lineAl:Rand}
\IF{$RAND < prDel$}
\STATE {\bf return} $true$
\ELSE
\STATE {\bf return} $false$
\ENDIF
\end{algorithmic}
\caption{\textit{delivery} function}\label{alg:DeliveryFunction}
\end{algorithm}

\subsection{A Permutation-based Path-relinking}\label{subsec:IPR&RestartStrategy}
To solve the VRPODTW, we apply two reinforcement procedures to \textcolor{black}{this} BRKGA-VM, which have been used in various metaheuristic and genetic algorithms (see, e.g., \citet{ref:brkga_mp_ipr}, \citet{ref:path_relinking_1}, and \citet{ref:path_relinking_2}).

The first one is the bidirectional Permutation-based \textcolor{black}{Implicit Path-Relinking (IPR-Per)}, reported \textcolor{black}{in Algorithm}~\ref{alg:IPR}. \textcolor{black}{The pseudocode for this procedure employs specific terms \textcolor{black}{and variables. For clarity}, a concise explanation of these is available in Table~\ref{table:glossary}.
IPR-Per receives as input two chromosomes, named \textcolor{black}{\textit{base} and \textit{guide}}, belonging to the hypercube $[0,1)^n$, and it returns a pair, denoted as \textcolor{black}{\textit{bestSolution}}, composed by the chromosome representing the best solution found and its fitness value. In general, a bidirectional path-relinking traces a path, defined as a set of intermediate solutions, in the problem space that connects two endpoint solutions, applying a series of moves that gradually transform one endpoint into the other.
In this case, these extremities are the solutions induced by the \textit{base} and \textit{guide} chromosomes. The algorithm alternates between examining the local variations around each of these two chromosomes. For the sake of clarity, \textcolor{black}{it is worth noting} that from this point forward, we will use the terms “solutions induced by chromosomes" and “chromosomes" interchangeably, treating them as essentially the same concept in our discussion.}

Lines~\ref{lineAl:start_init}--\ref{lineAl:end_init} initialize the elements of the procedure. In line 2, the parameter $pct_p$ sets the \textit{pathSize} which determines whether the path-relinking procedure covers the entire path between the two chromosomes or if it is truncated at a specific point.
The index list $RI$ is the set of gene positions on a generic chromosome.
Initially, this list is associated with the customer sub-chromosome (line~\ref{lineAl:RI_list}) and later it is associated with the driver sub-chromosome (line~\ref{lineAl:RI_change}).

\textcolor{black}{On the other hand, lists} $I_b$ and $I_g$ represent the sets of gene positions of \textit{base} and \textit{guide} chromosomes, respectively. For example, initially $I_b[1] = 1$ is the position of the first gene in the chromosome \textit{base}.
However, this position could change to any other value following the chromosome sorting process.

\textcolor{black}{In accordance with the structure of the decoder described in the previous section, the IPR-Per procedure also considers each of the aforementioned last two index lists as separated into two parts. The first one represents the set of customers, whereas \textcolor{black}{the second represents} the set of all drivers. In fact, in line~\ref{lineAl:base_sort} and~\ref{lineAl:guide_sort}, the indices corresponding to the customers \big($I_b[1:c], I_g[1:c]$\big), and drivers $\big(I_b[c+1:n], I_g[c+1:n]$\big) are sorted separately using the \textit{sort()} function based on the non-decreasing order of the genes in the \textit{base} and \textit{guide} chromosomes, respectively. For instance, if the first gene of the \textit{base} chromosome is the fourth smallest, then $I_b[4] = 1$.}

In lines~\ref{lineAl:start_bigFor}--\ref{lineAl:end_bigFor}, the path between the two chromosomes is built.
Specifically, in the loop~\ref{lineAl:start_RI_for}--\ref{lineAl:end_RI_for}, the procedure looks for the index that leads to the optimal gene swap in terms of solution quality for the chromosome $base$. 
In line~\ref{lineAl:bestInit} two variables are initialized. The first one, $i_{best}$, keeps track of the index that yields the best solution during the current iteration of the for loop. On the other hand, $val_{best}$ keeps track of the value of the best solution found during the iteration. Both variables are updated when a solution with a lower value is found, since we are considering a minimization problem (line~\ref{lineAl:bestUpdate}). If the positions of the corresponding genes in the \textit{base} and \textit{guide} chromosomes are the same, there is no change in the solution, therefore the algorithm removes that index from the \textit{RI} set and proceeds to the next iteration of the outer for loop (lines~\ref{lineAl:startIfSame}--\ref{lineAl:endIfSame}).

In line~\ref{lineAl:keys_swap}, a temporary swap is performed between two genes of the \textit{base} chromosome, specifically at the positions indicated by the index $i$ in the $I_b$ and $I_g$ lists.
This temporary swap serves as a trial to evaluate the solution cost of the new arrangement by running the decoder (line~\ref{lineAl:runDecoder}). Once the most advantageous swap is identified, it is then made permanent in line~\ref{lineAl:keys_swap_perma}. If this swap results in an improved solution, the $bestSolution$ is subsequently updated, as indicated in lines~\ref{lineAl:start_update}--\ref{lineAl:end_update}.
In line~\ref{lineAl:RI_update}, the index $i_{best}$, associated with the best swap, is eliminated from the index list $RI$. Then, in line~\ref{lineAl:chrom_swap}, the role of the \textit{base} and \textit{guide} solutions are exchanged, and this defines the bidirectional variant of path-relinking. Before starting a new iteration, the $pathSize$ is decreased by one. 

In each iteration of a standard path-relinking algorithm, the \textit{base} and \textit{guide} solutions progressively converge, resembling each other more closely. Typically, the procedure monitors the disparities between the two solutions by directly manipulating their constituent elements.
However, in the context of a BRKGA, solutions are derived from permutations of genes. Consequently, the path-relinking process tracks the differences between the two solutions using lists of permutation indices, necessitating the use of $I_b$ and $I_g$ for this purpose.
This approach is particularly well-suited for operating within the hypercube $[0,1)^n$, as opposed to working directly on the problem's solution space, as is common in traditional path-relinking methods. Therefore, this technique is categorized as implicit path-relinking.

Finally, the procedure terminates based on one of three stopping criteria: either the execution time limit is reached, the $pathSize$ is depleted, or all checks on the indices $j$ (line \ref{lineAl:start_bigFor}) have been exhausted. Additionally, once all indices corresponding to customers have been examined, the $RI$ list is updated to incorporate the indices associated with drivers, as delineated in (lines~\ref{lineAl:start_RI_swap}--\ref{lineAl:end_RI_swap}). Subsequently, the procedure repeats the loop starting at line~\ref{lineAl:start_bigFor}.

Given that the IPR-Per is based on swaps between two solutions, to make it more effective, it accepts two solutions only if they are sufficiently different. In particular, we use the \textcolor{black}{Kendall} tau rank distance to measure the difference between the potential \textit{base} and \textit{guide} chromosomes, and we consider the parameter $mp$ which indicates the minimum distance that the two chromosomes must respect. For more details about the entire procedure, the reader is \textcolor{black}{referred to \citet{ref:brkga_mp_ipr}}.

\begin{table}
\caption{Glossary of variables and terms in IPR-Per Procedure.}\label{table:glossary}\centering
\begin{tabular}{p{2.1cm}p{14cm}}
\hline\noalign{\smallskip}
\textcolor{black}{\textbf{Variable/Term}} & \textcolor{black}{\textbf{Brief explanation}} \\[0.5ex]
\noalign{\smallskip}\hline\noalign{\smallskip}
\textcolor{black}{IPR-Per} & \textcolor{black}{Bidirectional Permutation-based Implicit Path-Relinking (IPR-Per) procedure, an algorithm for exploring the solution space.} \\[0.5ex]
\hline
\textcolor{black}{\textit{base} chrom.} & \textcolor{black}{One of the two endpoint solutions used in the IPR-Per procedure.} \\[0.5ex]
\hline
\textcolor{black}{\textit{guide} chrom.} & \textcolor{black}{The other endpoint solution used in the IPR-Per procedure.} \\[0.5ex]
\hline
\textcolor{black}{\textit{bestSolution}} & \textcolor{black}{A pair consisting of the chromosome representing the best solution and its fitness value.} \\[0.5ex]
\hline
\textcolor{black}{path} & \textcolor{black}{The set of intermediate solutions generated during the path-relinking process.} \\[0.5ex]
\hline
\textcolor{black}{\textit{pathSize}} & \textcolor{black}{Determines the length of the path to be explored during the path-relinking procedure.} \\[0.5ex]
\hline
\textcolor{black}{\textit{RI}} & \textcolor{black}{List of gene positions still to be analyzed on a generic chromosome, initially associated with the customer sub-chromosome and later with the driver sub-chromosome.} \\[0.5ex]
\hline
\textcolor{black}{\(I_b, I_g\)} & \textcolor{black}{Lists representing the sets of gene positions for the \textit{base} and \textit{guide} chromosomes, respectively.} \\[0.5ex]
\hline
\textcolor{black}{\(i_{\text{best}}, val_{\text{best}}\)} & \textcolor{black}{Keeps track of the index yielding and the value of the best solution during the current iteration, respectively.} \\[0.5ex]
\hline
\textcolor{black}{Kendall tau rank dis.} & \textcolor{black}{A metric to measure the dissimilarity between two ranked lists, used for the \textit{base} and \textit{guide} chromosomes.} \\[0.5ex]
\hline
\textcolor{black}{\(mp\)} & \textcolor{black}{Minimum distance that the \textit{base} and \textit{guide} chromosomes must respect to be considered different.} \\[0.5ex]
\hline
\textcolor{black}{\(pct_p\)} & \textcolor{black}{Parameter used to set the \textit{pathSize}.} \\[0.5ex]
\hline
\textcolor{black}{\textit{solutionValue}} & \textcolor{black}{The fitness of a decoded chromosome, used to evaluate the quality of solutions.} \\[0.5ex]
\noalign{\smallskip}\hline\noalign{\smallskip}
\end{tabular}
\end{table}
\begin{algorithm}[h!]
\begin{algorithmic}[1]
\STATE {\bf input} $base$ and $guide$ chromosomes, $n$ (the dimension of the chromosomes), and $c$ (the number of customers)
\STATE {\bf set} $pathSize$ = $\lceil n\times pct_p\rceil$\label{lineAl:start_init}
\STATE {\bf set} $bestSolution$ = $(base,+\infty)$
\STATE {\bf set} $RI = [1, \ldots, c]$\label{lineAl:RI_list}
\STATE {\bf set} $(I_b, I_g) = ([1, \ldots, n], [1, \ldots, n])$\label{lineAl:end_init}
\STATE $I_b[1:c].sort(base)$, $I_b[c+1:n].sort(base)$\label{lineAl:base_sort}
\STATE $I_g[1:c].sort(guide)$, $I_g[c+1:n].sort(guide)$\label{lineAl:guide_sort}
\FOR{$j$ in $[1,\ldots,n]$}\label{lineAl:start_bigFor}
 \STATE $(i_{best}, val_{best}) = (-1, +\infty)$\label{lineAl:bestInit}
 \FOR{$i$ in $RI$}\label{lineAl:start_RI_for}
 \IF{$I_b[i] == I_g[i]$}\label{lineAl:startIfSame}
 \STATE $RI = RI\setminus \{i\}$
 \STATE break
 \ENDIF\label{lineAl:endIfSame}
 \STATE $swap(base[I_b[i]], base[I_g[i]])$\label{lineAl:keys_swap}
 \STATE $solutionValue = decoder(base)$\label{lineAl:runDecoder}
 \STATE $swap(base[I_b[i]], base[I_g[i]])$
 \IF{$solutionValue < val_{best}$}
 \STATE $(i_{best}, val_{best}) = (i, solutionValue)$\label{lineAl:bestUpdate}
 \ENDIF
 \ENDFOR\label{lineAl:end_RI_for}
 \IF{$i_{best}$ == $-1$}\label{lineAl:start_RI_swap}
 \STATE $RI = [c+1, \ldots, n]$\label{lineAl:RI_change}
 \STATE Go to line 8
 \ENDIF\label{lineAl:end_RI_swap}
 \STATE $swap(base[I_b[i_{best}]], base[I_g[i_{best}]])$\label{lineAl:keys_swap_perma}
 \IF{$val_{best} < bestSolution.value$}\label{lineAl:start_update}
 \STATE $bestSolution = (base, val_{ best})$
 \ENDIF\label{lineAl:end_update}
 \STATE $RI = RI\setminus \{i_{best}\}$\label{lineAl:RI_update}
 \STATE $swap(base, guide)$\label{lineAl:chrom_swap}
 \STATE $pathSize$ $-= 1$
 \IF{$pathSize == 0$ or time limit is reached}
 \STATE break
 \ENDIF
\ENDFOR\label{lineAl:end_bigFor}
\STATE {\bf return} $bestSolution$
\end{algorithmic}
\caption{IPR-Per procedure}\label{alg:IPR}
\end{algorithm}

\subsection{Restart Strategy and Variable Neighborhood Descent }

The second procedure employed to strengthen the proposed genetic algorithm is a restart strategy. If during evolution, the population is stuck and apparently does not move towards a better solution, we try to move to another zone of the solution space, as \textcolor{black}{proposed in \citet{ref:restartbrkga}}.

In particular, if after $h$ iterations the best current solution has not improved the algorithm discards all chromosomes, except the best one, and restarts using a new seed for the random-keys generator.

\textcolor{black}{In this study, we employed two versions \textcolor{black}{of BRKGA-VM to} address the problem. The multi-population and multi-parent variant that incorporates both the restart strategy and the path-relinking is called VM. Conversely, the version that, in addition to these procedures, also integrates a local search is referred to as VM$+$L.} In particular, we considered the VND proposed in \citet{ref:vnd} as a local search.
It is performed at each BRKGA-VM restart, before discarding the chromosomes, in an effort to find \textcolor{black}{an improving solution}. The neighborhood structures present in the VND are the following:
\begin{itemize}
 \item 2-Opt: removes two arcs in the same driver path or in two distinct paths, then reconnects the path(s) using the two removed arcs;
 \item Move Node: removes a customer $c$ from a driver path $pa$ and inserts it in a path $pa'$ in a feasible position;
 \item Swap Inter-Path: performs a swap between two customers of two different paths in the best feasible position;
 \item Swap Intra-Path: performs a swap between two customers of the same path in the best feasible position;
 \item New Path: initializes a new path for a company driver with a Move Node;
 \item New Path best: initializes a new path for a company driver with a Move Node, if the solution is improved.
\end{itemize}

\begin{figure}
 \centerline{\includegraphics[scale=0.7]{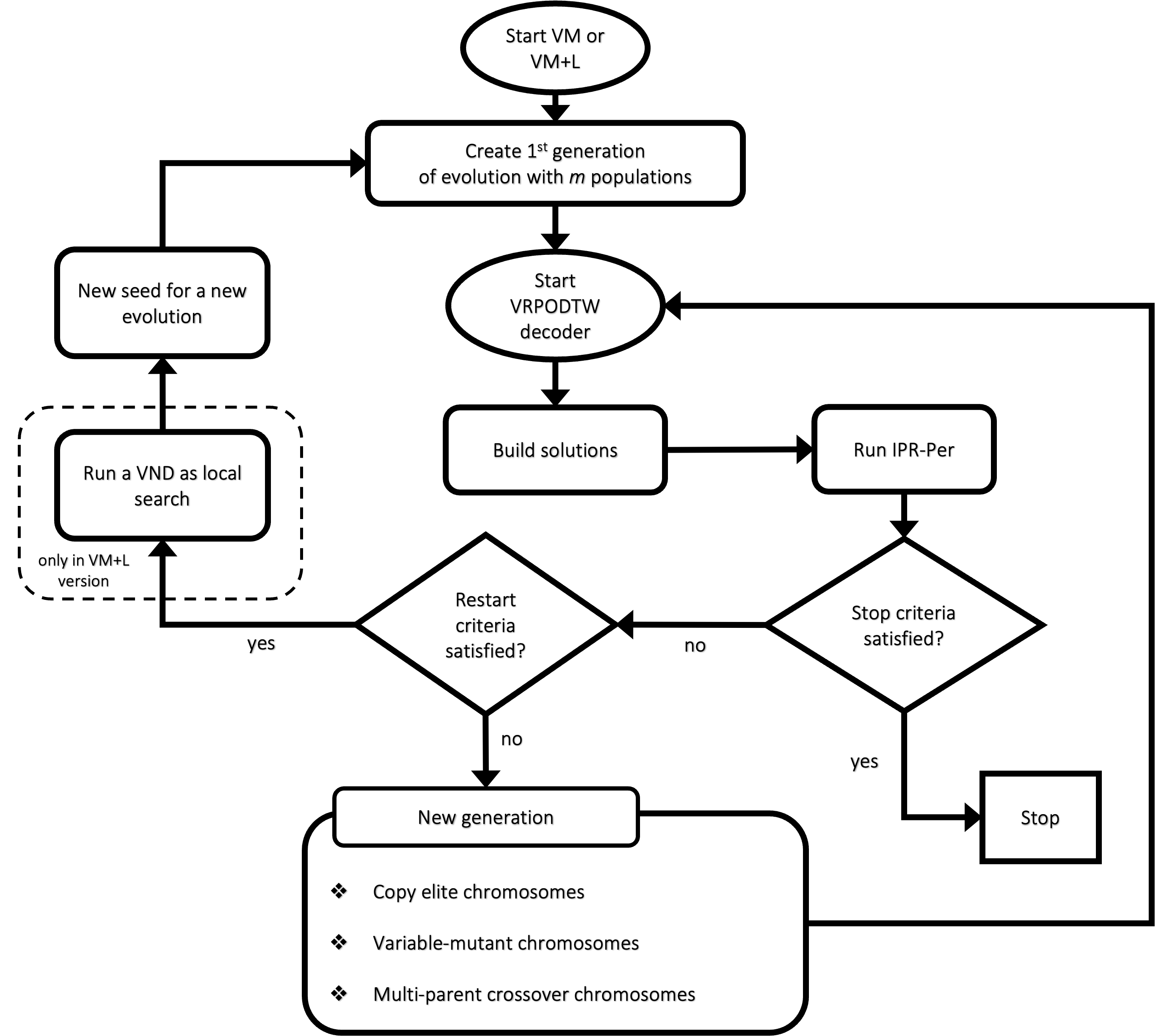}}
\caption{\textrm{Diagram of both metaheuristics considered to solve the VRPODTW.}}\label{fig:DiagramBRKGA}
\end{figure}

A summary of the main operations executed by the proposed metaheuristics \textcolor{black}{is shown in Figure~\ref{fig:DiagramBRKGA}}.
It starts by creating the first generation of $m$ populations and using a seed to generate all the chromosomes. In the second step the decoder converts the chromosomes in the VRPODTW solutions and consequently computes the fitness values, i.e., the total routing cost of the company vehicles plus the compensation for ODs. After decoding the chromosomes, the IPR-Per is performed to try to improve the best current solution receiving two chromosomes as input.
If both the \textcolor{black}{stopping} and restart strategy criteria are not reached, then the next step is to create a new generation with the three operators (copy repeat, variable-mutant and multi-parent crossover), and the process is repeated by decoding new populations.
If the restart criteria are reached, before discarding all chromosomes, except the best one, the VND is performed as a local search.
Regarding global \textcolor{black}{stopping} criteria, we consider two rules. The procedure is interrupted if either the set time limit or the maximum number of consecutive iterations without improvement ($wi$) are reached.

Table~\ref{tab:parameters} summarizes the parameters of the proposed BRKGA, grouping them into three sets: Operator; Path-relinking; Others.
Regarding the IPR-Per parameters, in addition to those already mentioned in the previous subsection, there is also the $sel$ parameter. It represents the method of choosing the \textit{guide} and \textit{base} chromosomes. \textcolor{black}{There are two possible options, $D_s:=\{randS, bestS\}$, which represent random selection and best chromosomes selection, respectively. In both cases the reference population is the elite one.}
The last column of the table includes the population size factor, the number of populations for each generation, the compensation factor $\rho$ for occasional drivers, the probability of delivery \textit{prDel}, and the restart parameter.

\section{Computational Study}\label{sec:ComputationalStudy}
In this section, we summarize the results of our computational experiments. 
The integration of the new BRKGA-VM variant has been implemented in C++. Considering the complexity due to numerous parameters, we performed preliminary tuning phases using the \textit{irace} package (refer to \citet{ref:irace} for details). This tool performs automatic configuration to optimize parameter values. The main code is written in R, while the interface between \textit{irace} and our metaheuristics has been developed in Python.

\textcolor{black}{To validate the innovative feature of variable mutant percentage of BRKGA introduced in this study, we conducted experimental tests with the VM algorithm and with the VM$+$L algorithm that integrates a local search into the procedure to improve the results. The outcomes were compared against two benchmarks. Firstly, they were analyzed in relation to the solutions obtained from the mathematical formulation discussed in \citet{ref:vrpodtw} for a set of small-sized instances. 
Secondly, for larger instances, they were compared with solutions generated by the multi-population and multi-parent BRKGA with IPR-Per and restart strategy, known as the MP algorithm and described in \citet{ref:vrp_brkga_3}, i.e., VM without the variable mutant percentage functionality.}

The mathematical model was coded in Java and solved to optimality using CPLEX 12.10, while the MP was coded in C++, using clang version 14.0.3. For the compilation, the C++17 standard was set using the CMAKE$\_$CXX$\_$STANDARD 17 specification in the CMake configuration file.
All the computational tests were conducted using a 2.6 GHz Intel Core i7-3720QM processor and 8 GB 1600 MHz DDR3 of RAM running macOs Catalina 10.15.7.

\subsection{Instances and Parameter Setting}\label{subsec:Instances}
Through a series of computational experiments, we evaluated the effectiveness of VM in solving the VRPODTW. We considered the instances used in \citet{ref:vrpodtw_vns}. We grouped them into four classes of increasing size.
Each class contains three different network typologies: clustered-type, random-type, and mixed-type. The first class (\textit{Test}) consists of a total of 36 instances, with 12 instances each for 5, 10, and 15 customers, respectively, each of the other three classes (\textit{Small}, \textit{Medium}, and \textit{Large}) consists of 15 instances with 25, 50, and 100 customers, respectively. In the instances, the number of occasional drivers is chosen in the set \{3, 5, 10, 15, 30\}, while the number of company drivers in the set \{3, 5, 8, 10\}. The instance characteristics are shown in Table~\ref{tab:InstanceCharacteristics}.
In particular, it reports the number of customers $\big(|C|\big)$, the interval to which the requests of each customer belong ($I_d$),
the number of company drivers $\big(|D|\big)$ and of occasional drivers $\big(|K|\big)$, the number of packages transportable by company vehicles ($Q$), and the interval to which the capacity of each OD vehicle belong ($I_{Q_k}$).
\begin{table}
\caption{Instance characteristics.}\label{tab:InstanceCharacteristics}
\centering
\begin{tabular}{llllll}
\hline
$|C|$ & $I_d$ & $|D|$ & $|K|$ & $Q$ & $I_{Q_k}$\\
\hline
5 & [4, 40] & 3 & 3 & 80 & [10, 25]\\
10 & [1, 40] & 3 & 3 & 80 & [10, 30]\\
15 & [1, 50] & 3 & 5 & 80 & [15, 35]\\
25 & [2, 40] & 5 & 10 & 100 & [20, 40]\\
50 & [1, 40] & 8 & 15 & 200 & [20, 40]\\
100 & [1, 50] & 10 & 30 & 400 & [20, 40]\\
\hline
\end{tabular}
\end{table}

We performed some tuning stages using the \textit{irace} program. Given that the only difference between the MP and VM metaheuristics lies in the variable functionality of the mutant population, to ensure a fair comparison between the two approaches and effectively evaluate the new version, we adopted the same parameter settings used in~\citet{ref:vrp_brkga_3} tuning with \textit{irace} only the two parameters in relation to the mutants, $pct^0_{vm}$ and $pct_{mi}$, as highlighted in Table~\ref{tab:TuningPhaseVM}.
Instead, for the VM$+$L, we tuned all parameters to be tuned: $m, \alpha$, $prDel$, \textit{Operator},  \textit{IPR-Per} (see Table~\ref{tab:TuningPhaseVML}).

\textcolor{black}{To capture the structural variation that occurs at different instance size scales and at the same time to avoid overfitting, we grouped the classes of instances into two sets for all \textit{irace} tuning phases; the first set (T\&S) contains the classes \textit{Test} and \textit{Small}, while the second contains \textit{Medium} and \textit{Large} classes. The parameters \textit{wi} and \textit{h} have been configured differently.
To further mitigate overfitting, ten randomly selected instances from each set were used to tune these parameters. Since the \textit{irace} extensively explores the parameter space, this can be not only more time-consuming but can also lead to excessive complexity in the optimization process. This is especially true if there is already a preliminary understanding of the values that might work better.}

In addition, for all experiments, we set $\rho$, the compensation factor for the occasional drivers (factor present in the objective function of the model), to 0.6, and we set the parameter \textit{wi} according to the size of the population of evolution, which we will detail in the next section.

For the setting of the restart parameter $h$, we examined the run time distribution in terms of iterations, taking into account $h\in A:=\{0, 100, 300, 500\}$, and we selected the best values.
We randomly selected three instances from each of the \textit{Small}, \textit{Medium}, and \textit{Large} classes and evaluated their run time distribution by summing the results over all runs for instances within the same class. This evaluation was conducted with consideration of the target values identified during a preliminary testing phase.
We would like to specify that on each instance, we performed 30 independent runs with different seeds. For this analysis the algorithm was stopped once a solution that is at least as good as the target was found.


\begin{figure}
\centering
\includegraphics[width=.75\textwidth]{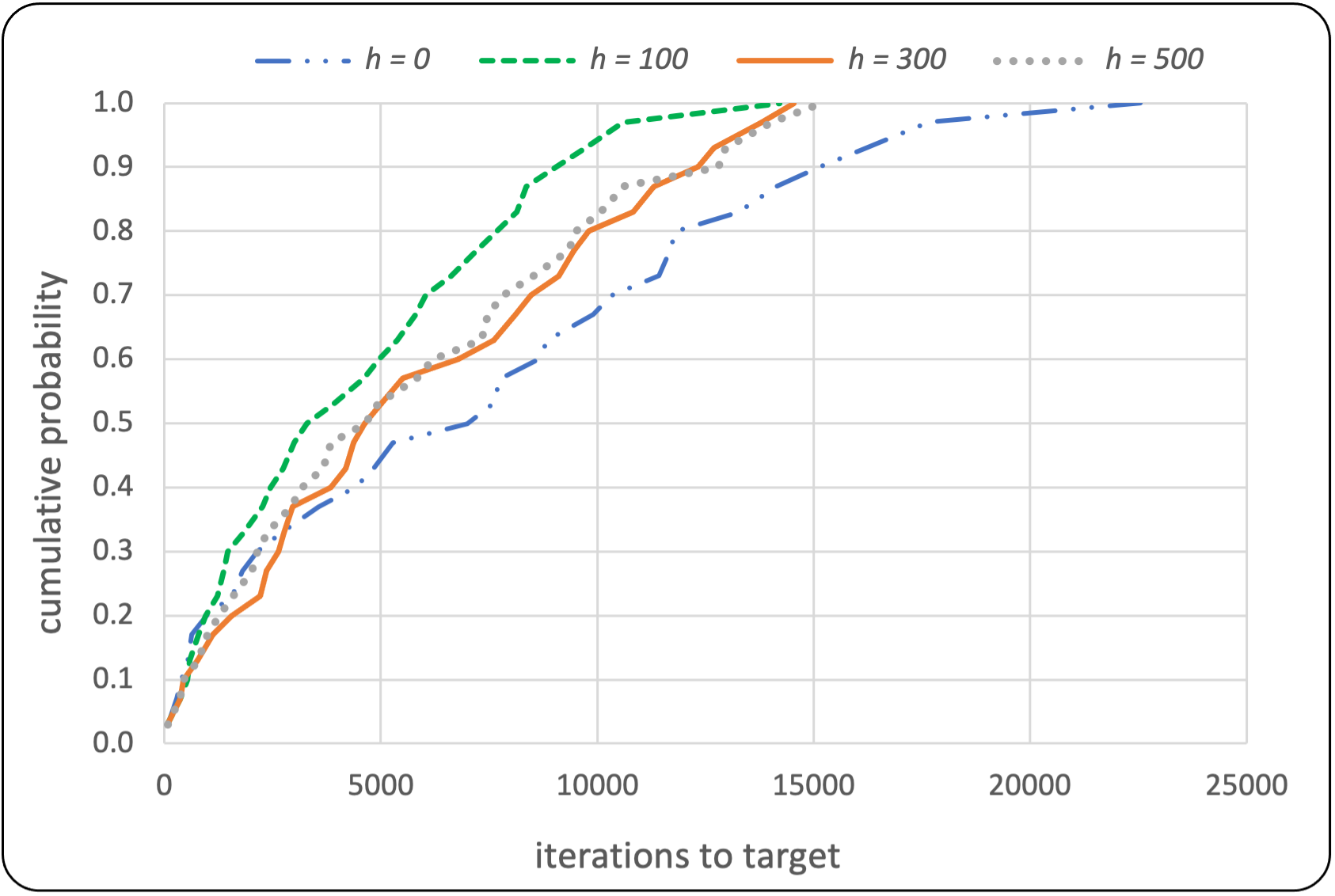}
\caption{Run time distribution for analyzing of the restart parameter $h$: Small class} \label{fig:IterationCount-small}
\end{figure}

\begin{figure}
\centering
\includegraphics[width=.75\textwidth]{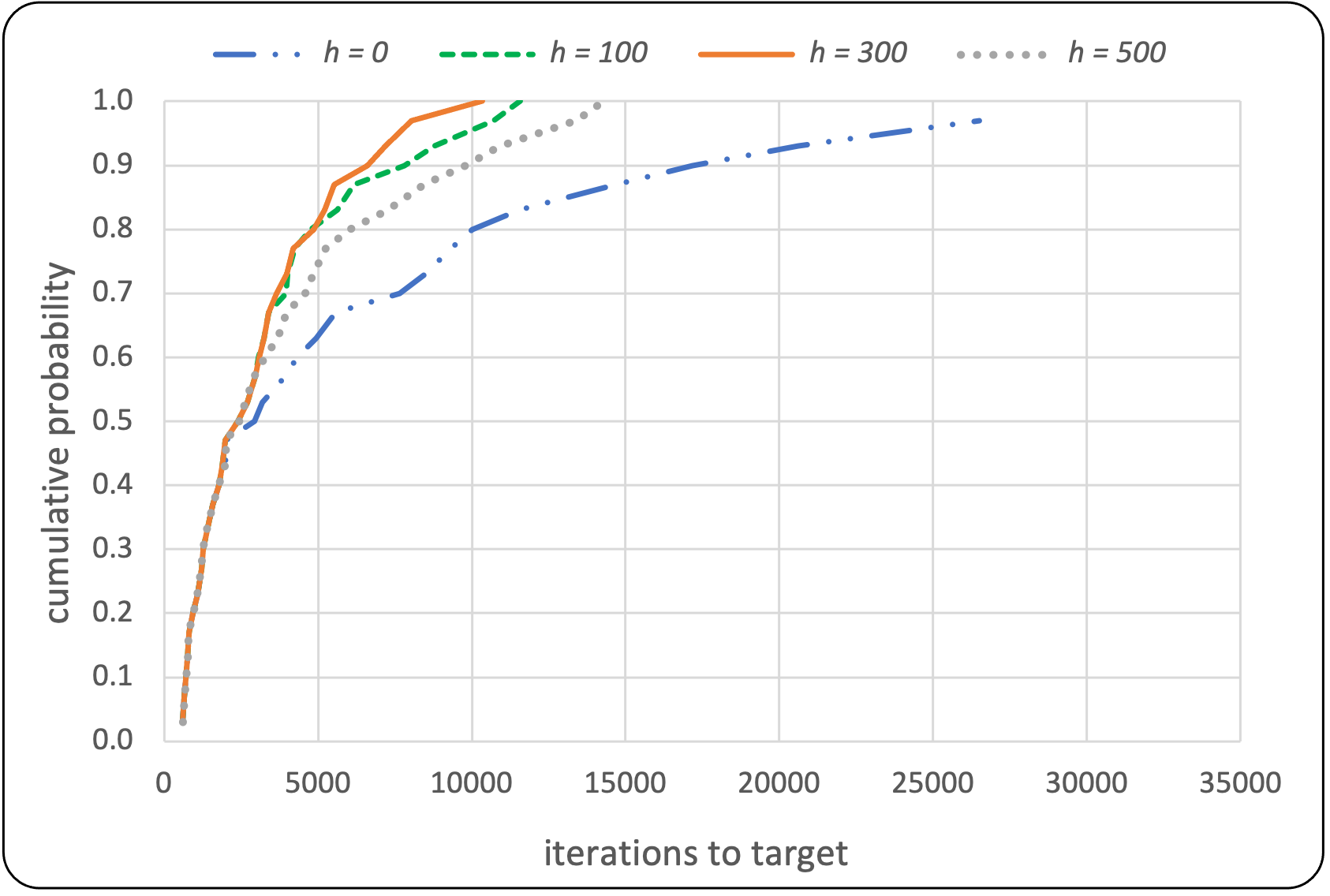}
\caption{Run time distribution for analyzing of the restart parameter $h$: Medium class} \label{fig:IterationCount-medium}
\end{figure}

\begin{figure}
\centering
\includegraphics[width=.75\textwidth]{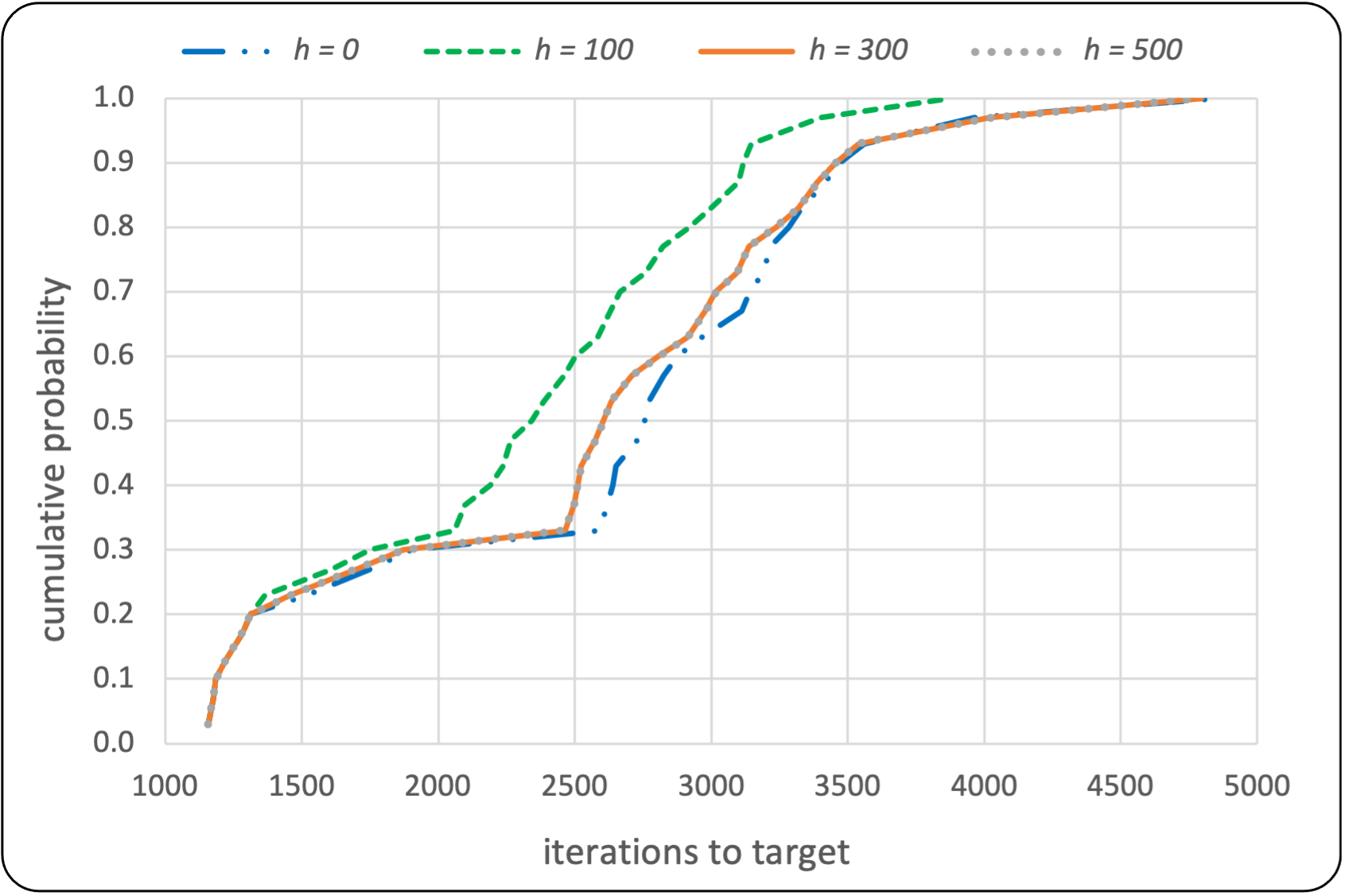}
\caption{Run time distribution for analyzing of the restart parameter $h$: Large class} \label{fig:IterationCount-large}
\end{figure}

Figs.~\ref{fig:IterationCount-small}~-- \ref{fig:IterationCount-large}
report the iteration count distribution plot for \textit{Small,
Medium}, and \textit{Large} classes, while
Table~\ref{tab:AnalysisRestartParameter} shows a summary of the
results of this analysis, in particular, it reports, for each class
and for each value of $h$, the minimum of upper bounds of iteration
counts for each quartile of the distribution (\textit{lower, median,
upper}), the maximum iteration count (\textit{max}), the average
iteration count (\textit{avg}), and its standard deviation
(\textit{sdev}).
As can be seen from Table~\ref{tab:AnalysisRestartParameter} and 
Figs.~\ref{fig:IterationCount-small}~-- \ref{fig:IterationCount-large},
in general, worst statistical indices are obtained when the parameter $h$ is set equal to 0, that is when the restart strategy is not applied.

For the \textit{Small} class the results are clearly in favor of a certain value. In fact, when $h$ is set to 100 we obtain, overall and for each instance, the quartiles, the average, and the standard deviation of number of iterations lower than those obtained by the other three parameter configurations.

In the \textit{Medium} class, observing the results and the plots, we deduce, in a more evident way than in the other two classes, that the restart strategy is not used for some runs. In fact, the distributions have similar lower tails and lower quartiles. The effect of the restart strategy begins to be observed from the median quartile onwards.
Overall and for each instance, with $h$ set to 300, lower statistical indices are obtained than those obtained with the other three values.

Finally, in the \textit{Large} class the advantage of the restart strategy is obtained only when $h$ is set to 100. With all the other values the distributions are very similar, and this happens because the size of the instances allows to reset the procedure a very few times.

As can be seen from Table~\ref{tab:AnalysisRestartParameter}, in particular, from standard deviations and averages, for each value of the parameter, the distribution in terms of iterations shows a clear decrease with increasing the size of classes, as expected. In other words, as the size of the instances increases, the skewed shape for the distributions decreases and the plots flatten towards the ordinate axis. This is due to the fact that, on average, the metaheuristic takes longer to do a single iteration.

%
\begin{table}
\caption{Summary of the analysis of the restart parameter. For the \textit{Small} class were considered the instances C103C25, C104C25, and RC102C25; for the \textit{Medium} class the instances C102C50, R105C50 RC104C50; for the \textit{Large} class the instances C105C100, 
R103C100, and RC104C100.}\label{tab:AnalysisRestartParameter}\centering
\begin{tabular}{p{1.4cm}p{0.7cm}p{1.1cm}p{1.2cm}p{1.1cm}p{1.1cm}p{1.5cm}p{1.3cm}}
\hline\noalign{\smallskip}
Class & $h$ & $lower$ & $median$ & $upper$ & $max$ & $avg$ & $sdev$ \\
\noalign{\smallskip}\hline\noalign{\smallskip}
$Small$	& 0 & 1806 & 6992 & 11682 & 22533 & 7437.23 & 5945.73\\
		& 100 & 1365 & 3303 & 7210 & 14217 & 4509.37 & 3566.84\\
		& 300 & 2361 & 4624 & 9451 & 14544 & 5989.93 & 4322.65\\
 		& 500 & 2025 & 4635 & 9331 & 15189 & 5766.43 & 4406.94\\
\noalign{\smallskip}\hline\noalign{\smallskip}
$Medium$& 0 & 1200 & 2926 & 9306 & 33062 & 6686.73 & 8105.43\\
		& 100 & 1200 & 2402 & 4226 & 11559 & 3395.50 & 2930.43\\
		& 300 & 1200 & 2402 & 4187 & 10338 & 3129.30 & 2424.65\\
 		& 500 & 1200 & 2440 & 5248 & 14284 & 4015.10 & 3797.76\\
\noalign{\smallskip}\hline\noalign{\smallskip}
$Large$	& 0 & 1747 & 2756 & 3218 & 4855 & 2615.87 & 941.26\\
 & 100 & 1607 & 2342 & 2823 & 3872 & 2287.50 & 747.80\\
		& 300 & 1698 & 2605 & 3137 & 4797 & 2552.93 & 929.00\\
 		& 500 & 1698 & 2605 & 3137 & 4797 & 2552.93 & 929.00\\
\noalign{\smallskip}\hline\noalign{\smallskip}
\end{tabular}
\end{table}
\begin{table}
\caption{Tuned VM algorithm parameters.\label{tab:TuningPhaseVM}}
\begin{tabular*}{\textwidth}{@{}@{\extracolsep{\fill}}llllllllllllll@{}}\hline
& \multicolumn{6}{l}{Operator} & \multicolumn{3}{l}{IPR-Per} & \multicolumn{4}{l}{Others} \\[-6pt]
& \multicolumn{6}{l}{\hrulefill} & \multicolumn{3}{l}{\hrulefill} & \multicolumn{4}{l@{}}{\hrulefill} \\
Parameter & $pct_e$ & $pct^0_{vm}$ & $pct_{mi}$ & $\pi_t$ & $\pi_e$ & $\phi$ & $sel$ & $md$ & $pct_p$ & $\alpha$ & $m$ & $prDel$ & $h$\\
\multicolumn{7}{l}{\hrulefill} & \multicolumn{3}{l}{\hrulefill} & \multicolumn{4}{l@{}}{\hrulefill} \\
Domain & - & \footnotesize{[0, 0.3]} & \footnotesize{[0, 0.3]} & - & - & - & - & - & - & - & - & - & -\\\hline
T\&S & 0.16 & 0.1 & 0.1 & 4 & 2 & $\frac{1}{r^2}$ & \footnotesize{randS} & 0.2 & 0.7 & 7 & 4 & 0.95 & 100\\
\textit{Medium} & 0.22 & 0.05 & 0.1 & 7 & 2 & $\frac{1}{r^2}$ & \footnotesize{randS} & 0.25 & 0.96 & 3 & 6 & 0.99 & 300\\
\textit{Large} & 0.22 & 0.05 & 0.1 & 7 & 2 & $\frac{1}{r^2}$ & \footnotesize{randS} & 0.25 & 0.96 & 3 & 6 & 0.99 & 100\\
\hline
\end{tabular*}
\end{table}
\begin{table}
\caption{Tuned VM$+$L algorithm parameters.\label{tab:TuningPhaseVML}}
\begin{tabular*}{\textwidth}{@{}@{\extracolsep{\fill}}llllllllllllll@{}}\hline
& \multicolumn{6}{l}{Operator} & \multicolumn{3}{l}{IPR-Per} & \multicolumn{4}{l}{Others} \\[-6pt]
& \multicolumn{6}{l}{\hrulefill} & \multicolumn{3}{l}{\hrulefill} & \multicolumn{4}{l@{}}{\hrulefill} \\
Parameter & $pct_e$ & $pct^0_{vm}$ & $pct_{mi}$ & $\pi_t$ & $\pi_e$ & $\phi$ & $sel$ & $md$ & $pct_p$ & $\alpha$ & $m$ & $prDel$ & $h$\\
\multicolumn{7}{l}{\hrulefill} & \multicolumn{3}{l}{\hrulefill} & \multicolumn{4}{l@{}}{\hrulefill} \\
Domain & \footnotesize{[0.1, 0.25]} & \footnotesize{[0, 0.3]} & \footnotesize{[0, 0.3]} & \multicolumn{2}{c}{\footnotesize{\{1, \ldots, 15\}}} & $D_\phi$ & $D_s$ & \multicolumn{2}{c}{\footnotesize{[0, 1]}} & \footnotesize{[1, 20]} & \footnotesize{\{1, \ldots, 5\}} & \footnotesize{[0, 1]} & -\\\hline
T\&S & 0.10 & 0.13 & 0.10 & 9 & 5 & $\frac{1}{r^2}$ & \footnotesize{randS} & 0.59 & 0.50 & 10 & 3 & 0.95 & 100\\
\textit{Medium} & 0.16 & 0.16 & 0.23 & 10 & 7 & $\frac{1}{r^2}$ & \footnotesize{randS} & 0.38 & 0.46 & 3 & 5 & 0.99 & 300\\
\textit{Large} & 0.16 & 0.16 & 0.23 & 10 & 7 & $\frac{1}{r^2}$ & \footnotesize{randS} & 0.38 & 0.46 & 3 & 5 & 0.99 & 100\\
\hline
\end{tabular*}
\end{table}

\subsection{Experimental Results}\label{subsec:Results}
In this section, we validate VM and VM$+$L examining and comparing the solutions obtained from the metaheuristic with known solutions or experimental data. To perform the first validation, we begin by comparing the two metaheuristics, the exact model, and the MP algorithm on the \textit{Test} class. Then, we analyse and investigate effectiveness and efficiency on the \textit{Small}, \textit{Medium}, and \textit{Large} classes.

\subsubsection{Experiments on Test Class}
To analyze the performance of VM and VM$+$L in solving the VRPODTW, in the first phase of the computational experiments, we make a comparison between the solutions found by the proposed metaheuristics and the optimal solutions determined by solving the mathematical model using CPLEX on the \textit{Test} class. In addition, to make a comparison with the known results in literature, we also included those of the MP algorithm. These results are reported in Table~\ref{tab:TestClass}, with each row presenting the averaged results (both time and cost) across 12 instances. Concerning the three metaheuristics under discussion, which are inherently non-deterministic, it should be noted that the solution costs and times for each instance were obtained as the average of 30 independent executions. These runs are conducted using distinct seeds for the chromosome generator to ensure variability and robustness of the resulting data. In the $Cost_{C}$ and $Time_{C}$ columns, there are the cost solutions and run times obtained with CPLEX. Whereas, the other columns display the cost solutions and run times for the MP, VM, and VM$+$L algorithms, respectively. For clarity, it is important to specify that all times reported in the table are in seconds. In the $Gap_{\%}$ columns there are the values of the average optimal gap on cost, calculated as: $Gap_{\%}:=100\cdot (Cost_{A}-Cost_{C})/Cost_{C}$, where A represents one of the three algorithms, MP, VM or VM$+$L.

For all three metaheuristics, we set the time limit equal to 900 seconds, and, in accordance with the size of the instances, we set the maximum of consecutive iterations without improvement \textit{wi} to 50, 750, and 2000 for instances with 5, 10, and 15 customers, respectively.
The pre-superscript numbers on the $Gap_\%$ columns indicates the number of instances that are not optimally solved within the two stop criteria.
The results show that VM and VM$+$L are very effective, in fact they achieve the optimal solutions in almost all instances in an efficient manner, producing reliable results within an acceptable time frame. \textcolor{black}{Specifically, the VM algorithm finds optimal solutions in all instances with 5 and 10 customers, while in the instances of with 15 customers, it achieves optimal solutions in all but three cases.} While, VM$+$L finds optimal solutions in all instances. Notably, on instances with 15 customers they find the solutions in an average time less than CPLEX. However, for instances with 5 and 10 customers the run time is not comparable: the optimization software finds the solutions much faster than all three metaheuristics.
In addition, Table~\ref{tab:TestClass} clearly shows that VM and VM$+$L are more effective and efficient than MP for this class of instances. Specifically, MP failed to find the optimal cost solution in 74 runs on three instances with 15 customers. This represents 80\% of the total number of runs performed on these instances. Moreover, regarding the execution time, VM and VM$+$L are, on average, faster that MP.

\begin{table}
\centering
\caption{Average results obtained on the instances belonging to the \textit{Test} class.}\label{tab:TestClass}
\begin{tabular}{llllllllllll}\hline
& \multicolumn{2}{l}{CPLEX} & \multicolumn{3}{l}{MP} & \multicolumn{3}{l}{VM} & \multicolumn{3}{l}{VM$+$L} \\[-6pt]
& \multicolumn{2}{l}{\hrulefill} & \multicolumn{3}{l@{}}{\hrulefill} & \multicolumn{3}{l@{}}{\hrulefill} & \multicolumn{3}{l}{\hrulefill} \\
$|C|$ & $Cost_C$ & $Time_C$ & $Cost$ & $Time$ & $Gap_{\%}$ & $Cost$ & $Time$ & $Gap_{\%}$ & $Cost$ & $Time$ & $Gap_{\%}$\\\hline
5 & 136.45 & 0.06 & 136.45 & 0.92 & 0.00\% & 136.45 & 0.59 & 0.00\% & 136.45 & 0.59 & 0.00\% \\
10 & 223.13 & 0.61 & 223.13 & 31.22 & 0.00\% & 223.13 & 18.55 & 0.00\% & 223.13 & 19.20 & 0.00\% \\
15 & 280.92 & 114.59 & 283.61 & 122.01 & $^3$0.81\% & 281.09 & 87.33 & $^3$0.07\% & 280.92 & 108.73 & 0.00\% \\
\hline
\end{tabular}
\end{table}

\subsubsection{Experiments on Small, Medium and Large Classes}
In a second phase of the experiment, we analyze the performance of VM and VM$+$L, in terms of effectiveness and efficiency, in comparison to MP on the \textit{Small}, \textit{Medium}, and \textit{Large} classes.

\textcolor{black}{Additionally, to obtain a lower bound for the solutions on instances of these three classes and to gain a deeper understanding of the quality of the results of the algorithms analyzed, experiments were also conducted using CPLEX. Despite CPLEX not being able to determine optimal solutions for all instances, it was crucial in determining the optimality gap. This gap represent the discrepancy between the best integer solution value found and the estimated optimal value of the problem, which was also calculated by the solver. This method proves to be useful when the solver reaches a time limit without finding the optimal solution. Indeed, if time expires but at least one feasible solution has been found, it returns the value of the best feasible solution and the related gap. In Tables~\ref{tab:SmallClass} and~\ref{tab:MediumClass}, in the $Cost_{C}$ and $Time_{C}$ columns, there are the cost solutions and run times obtained with CPLEX, whereas the gaps between the best solutions and the optimal values estimated by the solver are indicated in the $Gap_C\%$ column. If the solution is optimal, the $Gap_C\%$ column displays 0.00\%.
In the examination of the \textit{Large} class instances, the outcomes from the exact model are entirely disregarded. This is due to the inability of CPLEX to identify any feasible solutions within one-hour time limit for any of the instances under consideration.}
Referring to metaheuristics, Tables~\ref{tab:SmallClass}--\ref{tab:LargeClass} show the collected results averaged on 30 different runs. In these tables, in addition to the CPLEX information, the solution costs and run times of the three metaheuristics are also reported. The letter “C" in the test name stands for clustered-type instance; “R" for random-type, and “RC" for mixed-type. Also in these tables, all the times reported are calculated in seconds.

Table \ref{tab:RelGaps} documents the relative gap of the metaheuristics VM and VM$+$L compared to MP. This table presents the average percentage values of $RelGap_{\%}$ for all instance classes. Each row represents a specific test, while the columns display the results for VM and VM$+$L on the different instance classes.
For both metaheuristics, the relative gap was calculated using the following formula: $RelGap_{\%}:=100\cdot (Cost_{A}-Cost_{MP})/Cost_{MP}$, where A represents either of the two algorithms, VM or VM$+$L.
In the $AVG$ row we report the averages of the values in the column.

We set the time limit equal to 900 seconds, and \textit{wi} to 2500, 1500, and 1000 for instances with 25, 50, and 100 customers, respectively.

Regarding quality of the solutions, it is evident that on average, VM and VM$+$L perform better than MP; indeed, the average $RelGap\%$ values for all three classes of instances are less than zero (Table~\ref{tab:RelGaps}).
When we examined the gaps in depth, we saw a distinct pattern among the classes.
%
In the \textit{Small} class, while the VM algorithm outperforms MP in all instances, the VM$+$L algorithm does so in 13 (87\%) instances but with an overall greater relative gap than VM, (about $-3\%$ versus $-1.32\%$). In addition, VM performs better in three instances (C101C25, C104C25, and RC102C25).
VM$+$L is more effective than MP in all random-type and mixed-type instances. In the best case (R102C25), the relative gap value is about $-13\%$, while in the worst case (C104C25) MP outperforms VM$+$L, but the worsening is limited (i.e., a positive relative gap of $1.15\%$ is observed).
In this class, the effectiveness of both metaheuristics is much more pronounced in random instances and less evident in cluster and mixed instance types.
\textcolor{black}{On average, CPLEX takes about 40 minutes to identify \textcolor{black}{six} optimal solutions and in the remaining \textcolor{black}{nine} instances, within the established time limit, it finds feasible solutions with an average gap of 35\% compared to the theoretical optimum. In instances where CPLEX finds optimal solutions, the discrepancy with the solutions identified by the three metaheuristics is not significant. A noteworthy case is the instance RC101C25, where both VM and VM$+$L approaches, in all 30 independent runs, identify the optimal solution in an average time even shorter than that of the solver.}

The computational results obtained from the \textit{Medium} class reveal that, on average, the relative gap of VM solutions exhibits a slight improvement when compared to the corresponding results from MP for each instance type.
In contrast, VM$+$L solutions show a significant advantage in random-type instances compared to the results from the other two classes. In fact, the average $RelGap_{\%}$ for this particular instance type is approximately $-16\%$ for the \textit{Medium} class, while it stands at around $-8\%$ and $-6\%$ for the \textit{Small} and \textit{Large} classes, respectively. Moreover, in random-type instances, VM$+$L consistently outperforms MP with high solution quality. Notably, the lowest observed $RelGap_{\%}$ value is approximately $-12\%$ (R101C50).
Unlike the other two classes, VM$+$L finds better solutions than MP not only for the random-type instances, but for all instances of the class, and consequently, the overall average $RelGap_{\%}$ value, which is about $-8\%$, is the best among all the classes.
For this class as well, on average, VM$+$L finds better solutions than VM, except for four instances (C101C50, C104C50, RC104C50, and RC105C50).
\textcolor{black}{On average, CPLEX takes about 52 minutes to identify \textcolor{black}{two} optimal solutions and in the remaining 13 instances, within the established time limit, it finds feasible solutions with an average gap of 39\% compared to the theoretical optimum. In instances where CPLEX finds optimal solutions, the discrepancy with the solutions identified by the three metaheuristics is more significant compared to the first class.}

From the computational results obtained on the \textit{Large} class, it is evident that VM$+$L behaves worse than MP only on two instances, both of cluster-type, that is C102C100 (relative gap of $6.66\%$) and C105C100 (relative gap of $5.11\%$). Conversely, in these two cases, the VM proves to be the better approach among all three.
The highest relative gap across all experiments is achieved on the cluster-type instance C104C100, with a relative gap of $-31.05\%$.
Regarding the clustered-type and mixed-type instances, VM and VM$+$L are more effective in this class than in the other classes.
Furthermore, for VM$+$L there is more balance of effectiveness between the various types of instances. Conversely, for the VM algorithm, the instances of the random-type with 100 customers are where the smallest relative gap is achieved.
The value of the overall relative gap is approximately $-2\%$ and $-6\%$ for VM and VM$+$L respectively.

Regarding the computational overhead, the execution times of the three metaheuristics are comparable, VM$+$L takes, on average, a little more time to find the solutions, as we expected. This can be justified by the fact that with the integration of a local search a single iteration of the VM takes more time. The difference between the average running times decreases as the instance size increases, from a difference of about 85 seconds with the \textit{Small} class to a difference of only about 5 seconds with the \textit{Large} class.
Instead, the MP and VM algorithms have more similar times, and this leads us to confirm what we suspected: that varying the mutant population does not alter the algorithm's convergence speed.

Table~\ref{tab:lastSolutionUpdate} shows shows for the VM$+$L algorithm the number of runs for each class and type of instance, indicating whether or not local search was used to improve the solution. Since we performed 30 independent runs, with different seeds for each instance, the total number of runs for each instance type (cluster, random and mixed) amounts to 150. Column \textit{B} indicates the number of runs in which the best solution is found using only the genetic framework, meaning that the local search was not used to improve the initial solution. The other two columns indicate the number of runs in which both algorithms were used to improve the initial solution.
Specifically, when the last improvement of the solution, which led to the best solution, was carried out using local search, we denote it as $B+LS$. Conversely, when the last improvement was performed by VM$+$L, we denote it as $LS+B$.
On the basis of the considerations reported for the Tables~\ref{tab:SmallClass}--\ref{tab:LargeClass}, it is evident that the type of instances in which VM$+$L, on average, behaves the best is the random-type. This behaviour can be justified by considering the data reported in Table \ref{tab:lastSolutionUpdate}. In particular, the observed trend is due to a greater impact and involvement of local search to improve the solution in random instances. Another relevant fact that can be deduced from the comparison of the information reported in Table \ref{tab:lastSolutionUpdate} and the collected computational results is that the scenarios in which the effectiveness of VM$+$L is less evident are the cases in which the local search is involved less.
\begin{table}\caption{Computational results collected on the \textit{Small} class.}\label{tab:SmallClass}
\begin{tabular}{llllllllllll}\hline
& \multicolumn{3}{l}{CPLEX} & \multicolumn{2}{l}{MP} & \multicolumn{2}{l}{VM} & \multicolumn{2}{l}{VM$+$L} \\[-6pt]
& \multicolumn{3}{l}{\hrulefill} & \multicolumn{2}{l@{}}{\hrulefill} & \multicolumn{2}{l@{}}{\hrulefill} & \multicolumn{2}{l}{\hrulefill} \\
$Test$ & $Cost_C$ & $Time_C$ & $Gap_C\%$ & $Cost$ & $Time$ & $Cost$ & $Time$ & $Cost$ & $Time$ \\
\noalign{\smallskip}\hline\noalign{\smallskip}
C101C25  & 256.7 & 165.06 & 0.00\% & 258.16 & 415.40 &  257.08 &  511.07 & 257.08 & 520.40 \\
C102C25  & 272.2 & 3600 & 13.04\% & 280.37 & 375.56 &  277.56 &  442.47 & 281.95 & 485.85 \\
C103C25  & 258.8 & 3600 & 41.58\% & 262.09 & 433.05 &  262.07 &  381.05 & 261.88 & 547.03 \\
C104C25  & 266.3 & 3600 & 51.61\% & 257.30 & 386.79 &  257.27 &  335.91 & 260.26 & 501.14 \\
C105C25  & 286.0 & 3600 & 12.80\% & 288.57 & 476.53 &  287.29 &  315.59 & 286.93 & 436.66 \\
R101C25  & 318.7 & 1.29 & 0.00\% & 335.03 & 337.09 &  329.04 &  231.35 & 323.78 & 515.51 \\
R102C25  & 287.4 & 344.12 & 0.00\% & 333.55 & 305.81 &  324.68 &  314.91 &  289.77 & 450.24 \\
R103C25  & 303.7 & 941.60 & 0.00\% & 338.33 & 331.75 &  334.58 &  343.31 & 306.33 & 486.45 \\
R104C25  & 293.0 & 3600 & 6.81\% & 334.88 & 370.98 &  319.81 &  508.55 & 302.76 & 519.43 \\
R105C25  & 283.9 & 6.62 & 0.00\% & 317.72 & 420.67 &  305.57 &  405.63 & 302.33 & 439.36 \\
RC101C25 & 458.8 & 541.86 & 0.00\% & 459.10 & 442.15 & 458.80  & 350.16  & 458.80 & 501.00 \\
RC102C25 & 464.1 & 3600 & 43.46\% & 473.19 & 451.32 & 465.87  & 483.53  & 466.46 & 458.33 \\
RC103C25 & 437.0 & 3600 & 55.69\% & 449.21 & 433.89 & 443.85  & 467.80  & 440.34 & 474.54 \\
RC104C25 & 441.3 & 3600 & 53.15\% & 445.77 & 432.43 & 444.62  & 545.30  & 443.17 & 443.95 \\
RC105C25 & 543.6 & 3600 & 33.79\% & 537.15 & 369.77 & 532.25  & 407.57  & 531.05 & 477.64 \\
\noalign{\smallskip}\hline\noalign{\smallskip}
AVG	 & 344.77 & 2293.38 & 20.80\% & 358.03 & 398.88 & 353.36 & 402.95 & 347.23 & 483.37 \\
\noalign{\smallskip}\hline\noalign{\smallskip}
\end{tabular}
\end{table}
\begin{table}\caption{Computational results collected on the \textit{Medium} class.}\label{tab:MediumClass}
\begin{tabular}{llllllllllll}\hline
& \multicolumn{3}{l}{CPLEX} & \multicolumn{2}{l}{MP} & \multicolumn{2}{l}{VM} & \multicolumn{2}{l}{VM$+$L} \\[-6pt]
& \multicolumn{3}{l}{\hrulefill} & \multicolumn{2}{l@{}}{\hrulefill} & \multicolumn{2}{l@{}}{\hrulefill} & \multicolumn{2}{l}{\hrulefill} \\
$Test$ & $Cost_C$ & $Time_C$ & $Gap_C\%$ & $Cost$ & $Time$ & $Cost$ & $Time$ & $Cost$ & $Time$ \\
\noalign{\smallskip}\hline\noalign{\smallskip}
C101C50  & 341.1 & 3600 & 0.12\% & 407.06 & 485.89 &  402.36 &  452.32 & 405.40 & 513.71 \\
C102C50  & 312.5 & 3600 & 37.20\% & 418.70 & 455.05 &  421.06 &  428.42 & 374.64 & 478.05 \\
C103C50  & 427.9 & 3600 & 62.56\% & 458.83 & 381.86 &  452.16 &  395.99 & 387.13 & 567.81 \\
C104C50  & 481.3 & 3600 & 69.42\% & 381.15 & 537.35 &  377.36 &  510.00 & 380.38 & 342.27 \\
C105C50  & 331.2 & 3600 & 22.95\% & 378.71 & 500.48 &  378.03 &  459.58 & 376.49 & 518.67 \\
R101C50  & 535.3 & 250.08 & 0.00\% & 680.16 & 379.81 &  660.50 &  474.90 & 598.63 & 384.56 \\
R102C50  & 499.0 & 3600 & 11.07\% & 629.02 & 398.74 &  620.29 &  391.00 & 537.16 & 439.74 \\
R103C50  & 448.6 & 3600 & 32.52\% & 567.55 & 301.15 &  545.81 &  310.53 & 459.55 & 379.86 \\
R104C50  & 430.5 & 3600 & 34.16\% & 590.90 & 483.83 &  575.85 &  405.77 & 458.30 & 517.03 \\
R105C50  & 459.6 & 360.20 & 0.00\% & 625.66 & 557.50 &  602.50 &  440.09 & 543.98 & 512.37 \\
RC101C50 & 497.5 & 3600 & 7.68\% & 601.74 & 467.42 & 593.15  & 399.11  & 592.04 & 507.22 \\
RC102C50 & 549.8 & 3600 & 37.45\% & 584.02 & 476.36 & 589.16  & 669.12  & 577.95 & 477.67 \\
RC103C50 & 545.6 & 3600 & 60.10\% & 545.29 & 547.35 & 541.40  & 420.86  & 536.99 & 536.18 \\
RC104C50 & 532.5 & 3600 & 65.23\% & 535.62 & 428.10 & 514.67  & 661.07  & 515.77 & 433.85 \\
RC105C50 & 543.8 & 3600 & 57.41\% & 559.77 & 511.20 & 551.18  & 407.92  & 557.25 & 505.10 \\
\noalign{\smallskip}\hline\noalign{\smallskip}
AVG	 & 462.41 & 3160.69 & 39.24\% & 530.95 & 460.81 & 521.70 & 455.11 & 486.78 & 474.27 \\
\noalign{\smallskip}\hline\noalign{\smallskip}
\end{tabular}
\end{table}
\begin{table}\caption{Computational results collected on the \textit{Large} class.}\label{tab:LargeClass}
\centering
\begin{tabular}{lllllllll}\hline
& \multicolumn{2}{l}{MP} & \multicolumn{2}{l}{VM} & \multicolumn{2}{l}{VM$+$L} \\[-6pt]
& \multicolumn{2}{l@{}}{\hrulefill} & \multicolumn{2}{l@{}}{\hrulefill} & \multicolumn{2}{l}{\hrulefill} \\
$Test$ & $Cost_C$ & $Time_C$ & $Cost$ & $Time$ & $Cost$ & $Time$ \\
\noalign{\smallskip}\hline\noalign{\smallskip}
C101C100 & 1265.58 & 736.18 &  1215.16 & 769.30  & 1213.39 & 616.23 \\
C102C100 & 1244.17 & 522.76 &  1226.63 & 663.12  & 1327.02 & 607.49 \\
C103C100 & 1162.51 & 864.12 &  1158.53 & 640.73  & 1044.35 & 817.06 \\
C104C100 & 1128.94 & 704.22 &  1089.98 & 603.97  & 778.36 & 706.75 \\
C105C100 & 1246.46 & 495.77 &  1227.67 & 578.19  & 1310.16 & 612.20 \\
R101C100 & 1273.47 & 802.34 &  1267.49 & 900.00  & 1250.64 & 657.77 \\
R102C100 & 1231.16 & 685.67 &  1226.81 & 537.58  & 1209.19 & 680.67 \\
R103C100 & 1144.23 & 664.41 &  1131.96 & 605.23  & 1104.79 & 687.76 \\
R104C100 & 1082.31 & 514.60 &  1043.89 & 900.00  & 886.26 & 838.49 \\
R105C100 & 1184.51 & 708.55 &  1228.36 & 748.55  & 1139.06 & 895.81 \\
RC101C100 & 1241.41 & 784.90 & 1209.49 & 751.62 & 1208.69 & 806.41 \\
RC102C100 & 1274.27 & 716.03 & 1272.71 & 707.92 & 1217.31 & 833.15 \\
RC103C100 & 1156.43 & 502.74 & 1100.52 & 446.82 & 1091.51 & 606.96 \\
RC104C100 & 1127.42 & 900.00 & 1075.11 & 900.00 & 1040.14 & 583.10 \\
RC105C100 & 1236.53 & 883.79 & 1193.09 & 718.95 & 1174.27 & 608.92 \\
\noalign{\smallskip}\hline\noalign{\smallskip}	
AVG	 & 1199.96 & 699.07 & 1177.83 & 698.13 & 1133.01 & 703.92 \\
\noalign{\smallskip}\hline\noalign{\smallskip}
\end{tabular}
\end{table}
\begin{table}\caption{Average $RelGap_{\%}$ respect to MP
for all classes of instances.}\label{tab:RelGaps}
\centering
\begin{tabular}{lllllll}\hline
& \multicolumn{2}{l}{\textit{Small} class} & \multicolumn{2}{l}{\textit{Medium} class} & \multicolumn{2}{l}{\textit{Large} class}\\[-6pt]
& \multicolumn{2}{l}{\hrulefill} & \multicolumn{2}{l}{\hrulefill} & \multicolumn{2}{l}{\hrulefill} \\
$Test$ & $VM$ & VM$+$L & $VM$ & VM$+$L & $VM$ & VM$+$L \\
\noalign{\smallskip}\hline\noalign{\smallskip}
C101  & $-0.42$ & $-0.42$ &  $-1.15$ & $-0.41$ & $-3.98$ & $-4.12$ \\
C102  & $-1.00$ & $0.56$ &  $0.56$ & $-10.52$ & $-1.41$ & $6.66$ \\
C103  & $-0.01$ & $-0.08$ &  $-1.45$ & $-15.63$ & $-0.34$ & $-10.16$ \\
C104  & $-0.01$ & $1.15$ &   $-0.99$ & $-0.20 $ & $-3.45$ & $-31.05$ \\
C105  & $-0.44$ & $-0.57$ &  $-0.18$ & $-0.59 $ & $-1.51$ & $5.11$ \\
R101  & $-1.79$ & $-3.36$ &  $-2.89$ & $-11.99$ & $-0.47$ & $-1.79$ \\
R102  & $-2.66$ & $-13.13$ & $-1.39$ & $-14.60$ & $-0.35$ & $-1.78$ \\
R103  & $-1.11$ & $-9.46$ &  $-3.83$ & $-19.03$ & $-1.07$ & $-3.45$ \\
R104  & $-4.50$ & $-9.59$ &  $-2.55$ & $-22.44$ & $-3.55$ & $-18.11$ \\
R105  & $-3.82$ & $-4.84$ &  $-3.70$ & $-13.06$ & $3.70$ & $-3.84$ \\
RC101 & $-0.07$ & $-0.07$ &  $-1.43$ &  $-1.61$ & $-2.57$ &  $-2.64$ \\
RC102 & $-1.55$ & $-1.42$ & $0.88 $  & $-1.04$ & $-0.12$ &  $-4.47$ \\
RC103 & $-1.19$ & $-1.97$ &  $-0.71$ &  $-1.52$ & $-4.83$ &  $-5.61$ \\
RC104 & $-0.26$ & $-0.58$ &  $-3.91$ &  $-3.71$ & $-4.64$ &  $-7.74$ \\
RC105 & $-0.91$ & $-1.14$ &  $-1.53$ &  $-0.45$ & $-3.51$ &  $-5.04$ \\
\noalign{\smallskip}\hline\noalign{\smallskip}
AVG	 & $-1.32$ & $-3.08$ & $-1.62$ & $-7.79$ & $-1.87$ & $-5.87$ \\
\noalign{\smallskip}\hline\noalign{\smallskip}
\end{tabular}
\end{table}
\begin{table}
\caption{Number of local search execution in the last solution update.}\label{tab:lastSolutionUpdate}\centering
\begin{tabular}{lllll}
\hline\noalign{\smallskip}
Class & $type$ & $B$ & $B+LS$ & $LS+B$\\
\noalign{\smallskip}\hline\noalign{\smallskip}
 	& C & 86 & 16 & 48\\
	$Small$	& R & 10 & 91 & 49\\
 		& RC & 94 & 18 & 38\\
\noalign{\smallskip}\hline\noalign{\smallskip}
 & C & 61 & 26 & 63 \\
$Medium$ & R & 11 & 122 & 17\\
	 	 & RC & 103 & 2 & 45\\
\noalign{\smallskip}\hline\noalign{\smallskip}
	 & C & 58 & 18 & 74 \\
 $Large$ & R & 55 & 39 & 56\\
		 & RC & 9 & 94 & 47\\
\noalign{\smallskip}\hline\noalign{\smallskip}
\end{tabular}
\end{table}

\begin{figure}
\centering
\includegraphics[width=.45\textwidth]{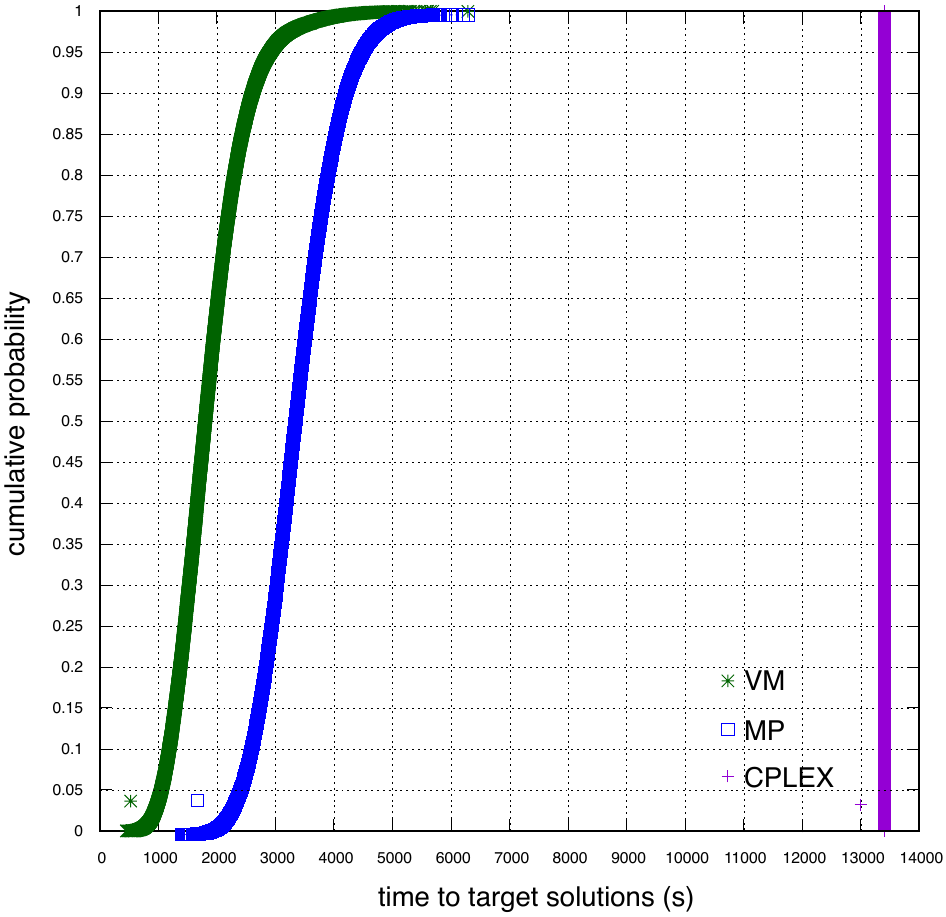}
\caption{Multiple time-to-target plots for VM, MP and CPLEX for all instances obtained by simulation with $6\cdot 10^4$ points each: Small class.} \label{fig:mttt_plots-small}
\end{figure}

\begin{figure}
\centering
\includegraphics[width=.45\textwidth]{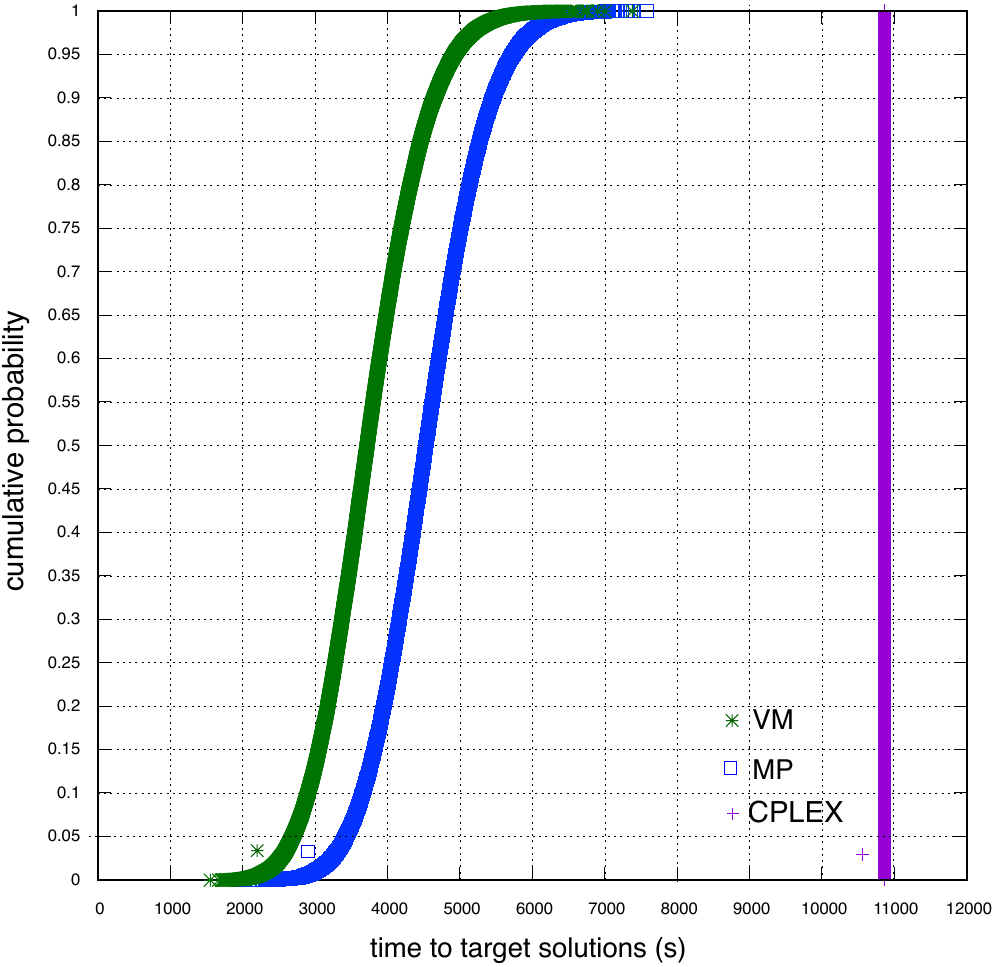}
\caption{Multiple time-to-target plots for VM, MP and CPLEX for all instances obtained by simulation with $6\cdot 10^4$ points each: Medium class.} \label{fig:mttt_plots-medium}
\end{figure}

\begin{figure}
\centering
\includegraphics[width=.45\textwidth]{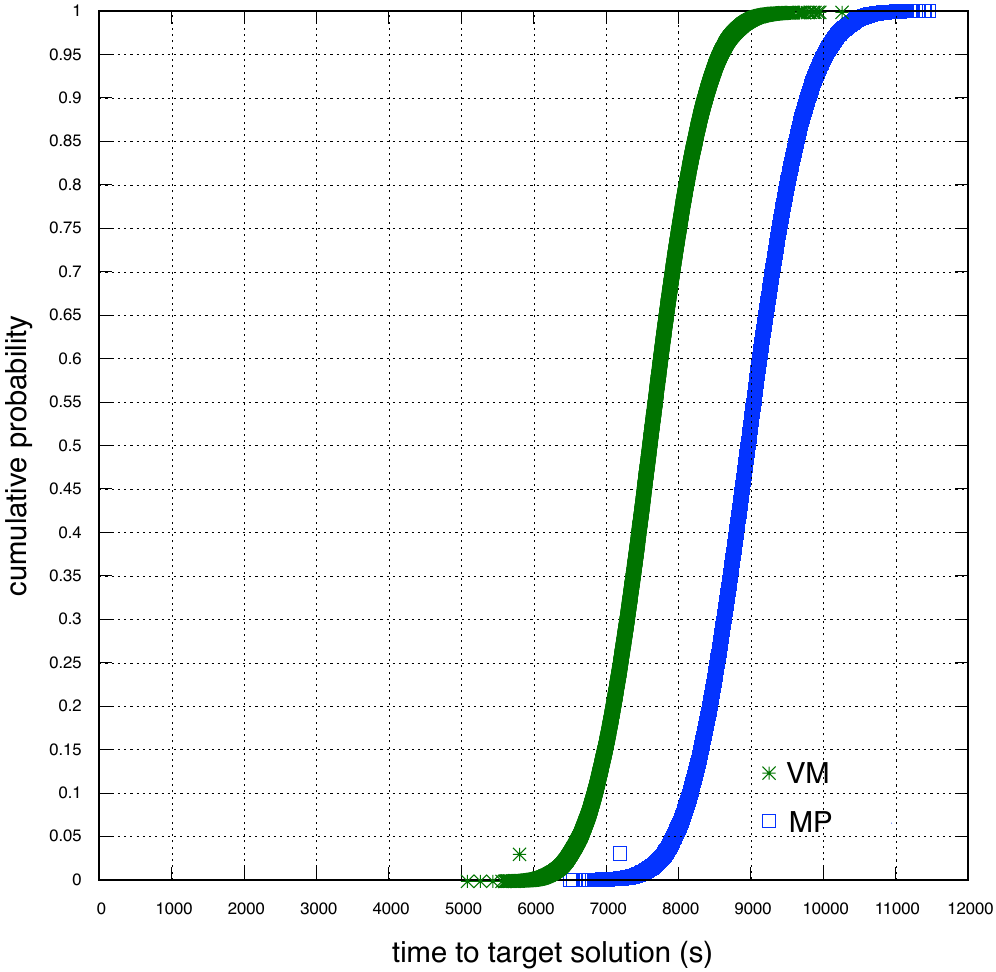}
\caption{Multiple time-to-target plots for VM, MP and CPLEX for all instances obtained by simulation with $6\cdot 10^4$ points each: Large class.} \label{fig:mttt_plots-large}
\end{figure}

\textcolor{black}{To enhance the analysis of the timing for the
approaches used and to go beyond the mere comparison of results
achieved within the predefined stopping criteria, we have conducted
a time-to-target test. We aim to scrutinize and compare the performance
of algorithms using multiple time-to-target plots (mttt-plots) as
described in~\citet{ref:mttt_plots}. Specifically, we seek to compare
the VM and the MP to underscore the efficacy of using a variable
mutant population in accelerating convergence to target solutions.
Furthermore, we aspire to compare the performance of the two
metaheuristics with that of CPLEX, given that in their prior
comparison, performance was assessed under varying time limit. This
allows for a more holistic and rigorous evaluation of the computational
efficiency of the techniques under review.
In general, this tool is useful for studying the convergence speed of algorithms to given values. For every class, all instances were considered, establishing a specific target value for each. These target values were deliberately chosen to be realistically achievable by each algorithm tested within a maximum limit of one hour. Specifically, for the two metaheuristics under examination, 60 independent runs were conducted, aiming to match or exceed the pre-established target value. 
Figures~\ref{fig:mttt_plots-small}~--
\ref{fig:mttt_plots-large}
illustrate the results of these tests through mttt-plots, developed
based on a sample of $6\cdot 10^4$ data points and aggregating the
run time for all instances. In the plot in Figure~\ref{fig:mttt_plots-large},
the data related to CPLEX are omitted because, as highlighted in
previous comparative analyses, it has not demonstrated the ability
to generate feasible solutions within the one-hour time limit.
In the graphs, the results of CPLEX are represented as vertical
lines, as we expected, since the algorithm is deterministic, but
we decided to show them to highlight the clear difference with the
times related to the metaheuristics. Considering what has just been
discussed, it makes sense for CPLEX to compare the time to achieve
target solutions with those attained by metaheuristics, specifically
concerning the certain achievement (the cumulative probability of
1) of target values.
In the \textit{Small} class, each metaheuristic certainly achieves the targets for all instances in about 2 hours, while CPLEX achieves them in almost 4 hours. A similar trend is also observed in the \textit{Medium} class, however with a smaller difference. In this class, each metaheuristic reaches the target values for all instances in about 2 hours, while CPLEX takes almost 3 hours. The one-hour difference between the two classes can be attributed to the fact that, in the class with 25 customers, there is an additional instance that is particularly difficult for CPLEX to solve in this test with target values.
Regarding metaheuristics, in the \textit{Small} class, on average, in about 200 seconds the probability of reaching the target value in a single instance for the MP is about 30\%, while in the same time the probability of the VM reaches even 95\%. Furthermore, to be 99\% confident of identifying the target values for all instances, a total execution time of 4892.57 seconds is required for the MP algorithm, and 3762.46 seconds for the VM. The application of the tttplots-compare tool, developed by\cite{ref:tttplots_compare}, reveals that the VM method outperforms MP for this set of 15 small instances and 15 targets. This finding is based on the fact that the tool has calculated that there is a 95\% probability that VM will take equal or less time than MP to achieve a solution of at least as good as the target.
In the \textit{Medium} class, on average, within roughly 300 seconds, the probability of hitting the target value in a single instance for MP is about 50\%, while in the same time, the probability for VM goes up to 90\%. In the case of medium instances, the tttplots-compare tool identifies a decrease in the difference between the performances of the two metaheuristics, but still maintaining a clear gap. In fact, in the second class, the tool has calculated that there is a 80\% probability that VM will take equal or less time than MP to achieve a solution of at least as good as the target. 
Finally, in the \textit{Large} class, on average, in about 570 seconds the probability of reaching the target value in a single instance for the MP is about 25\%, while in the same time the probability of the VM reaches even 95\%. Furthermore, to be 99\% confident of identifying the target values for all instances, a total execution time of 10458.04 seconds is required for the MP algorithm, and 8962.55 seconds for the VM. In the case of large instances, the tttplots-compare tool identifies a difference between the performances of the two metaheuristics similar to that of the first class. In fact, in the third class, the tool has calculated that there is a 94\% probability that VM will take a time equal to or less than MP to reach a solution at least equal to that of the objective.}

\section{Conclusions}\label{sec:conclusions}
In this paper, we presented an innovative variant of the biased random-key genetic algorithm, with variable mutant population (BRKGA-VM).

\textcolor{black}{In addition, to solve the vehicle routing problem with occasional drivers and time windows, we considered two metaheuristics based on the proposed new BRKGA variant. The first is a multi-population and multi-parent BRKGA-VM with IPR-Per and restart strategy, named VM. The second is a hybrid approach that integrates the former with a local search, named VM$+$L.}

We developed a decoding process that transforms chromosomes into driver paths. This process involves verifying the compatibility of time windows and travel times, as well as adding a random step that can be replicated for the same chromosome in subsequent generations.

To assess the performance of the proposed solution approach in terms of effectiveness and efficiency an extensive computational study was conducted considering instances of increasing size. 

\textcolor{black}{To demonstrate the efficacy of the variable mutant population, we compared the VM algorithm with MP, i.e. the multi-population and multi-parent BRKGA with IPR-Per and restart strategy, that is VM without the variable mutant percentage option.
In the comparison between MP and VM, we highlighted the effectiveness of parameter variability by analyzing Table~\ref{tab:RelGaps}. Across various instances, for each class, an increase in solution efficacy was observed using VM compared to MP, with varying percentages. These improvements, found in 93\% of the instances and in all types (clustered, random, and mixed), stem from the adaptability of the new version to different problems. Different challenges require varying levels of exploration and intensification, and the use of a variable percentage of mutants makes the algorithm more flexible, adapting to a wide range of problems without the need for manual adjustments.}
\textcolor{black}{Furthermore, as observed in 
Figures~\ref{fig:mttt_plots-small}~--
\ref{fig:mttt_plots-large},
VM reaches predetermined targets more quickly than MP. This outcome
is also influenced by MP's tendency to stagnate; increasing the
mutants induces greater variability in the population, facilitating
an escape from the stagnation area and exploring new regions of the
solution space. This observation, combined with the fact that the
new version of the genetic algorithm does not improve the rate of
convergence, leads us to conclude that, although the VM allows for
obtaining good solutions more quickly compared to those previously
known in the literature, it still requires a significant amount of
time to attempt further improvements of the solutions already found.}
s
Furthermore, the conducted analysis revealed a positive impact of local search in terms of solution quality, especially for random-types instances, in fact, the results showed that the proposed VM$+$L is more effective on average than the other metaheuristic considered for comparison.

\section*{Acknowledgments}
This work is supported by the Italian Ministry of University Education and Research (MIUR), project: ``Innovative approaches for distribution logistics" - Code DOT1305451 - CUP H28D2000002\-006.



\bibliographystyle{plainnat}
\bibliography{itor}
\end{document}